\documentclass[11pt]{article}

\PassOptionsToPackage{table}{xcolor}
\usepackage[final]{acl}
\usepackage{times}
\usepackage{latexsym}

\usepackage[T1]{fontenc}

\usepackage[utf8]{inputenc}

\usepackage{microtype}

\usepackage{inconsolata}

\usepackage{graphicx}
\usepackage{booktabs}
\usepackage{amsmath}
\usepackage{amssymb}
\usepackage{subcaption}
\usepackage[most]{tcolorbox}
\usepackage{enumitem}
\usepackage{multirow}
\usepackage{array}
\usepackage{tabularx}
\usepackage{makecell}
\usepackage{listings}

\lstdefinestyle{promptstyle}{
  basicstyle=\ttfamily\scriptsize,
  breaklines=true,
  breakatwhitespace=false,
  columns=fullflexible,
  keepspaces=true,
  showstringspaces=false,
  upquote=true,
  tabsize=2
}

\newcommand{\stdpart}[1]{%
  \kern0.04em%
  {\normalfont\normalcolor
   \fontsize{5.5}{6.0}\selectfont
   \raisebox{0.08ex}{\ensuremath{\pm}#1}}%
}

\newcommand{\meanstd}[2]{%
  \mbox{#1\stdpart{#2}}%
}

\newcommand{\bestmeanstd}[2]{%
  \mbox{\textbf{#1}\stdpart{#2}}%
}

\newcommand{\secondmeanstd}[2]{%
  \mbox{\underline{#1}\stdpart{#2}}%
}

\definecolor{topone}{RGB}{255,235,235}
\definecolor{toptwo}{RGB}{235,245,255}

\newcolumntype{Y}{>{\centering\arraybackslash}X}
\title{ChJGBench: A Comprehensive Benchmark for Chinese Legal Judgment Generation with Diagnostic Evaluation}

\author{
    \textbf{Yinghao Hu}\textsuperscript{\normalfont 1},
    \textbf{Linggang Kong}\textsuperscript{\normalfont 2},
    \textbf{Qian Xiao}\textsuperscript{\normalfont 3},
    \textbf{Yuting Huang}\textsuperscript{\normalfont 1},
    \textbf{Yunlong Zhang}\textsuperscript{\normalfont 2},
    \\
    \textbf{Yujie Zhang}\textsuperscript{\normalfont 1},
    \textbf{Yiquan Wu}\textsuperscript{\normalfont 1,*},
    \textbf{Fei Duan}\textsuperscript{\normalfont 3},
    \textbf{Kaisong Song}\textsuperscript{\normalfont 3,4},
    \\
    \textbf{Jun Lin}\textsuperscript{\normalfont 3},
    \textbf{Kun Kuang}\textsuperscript{\normalfont 1,*},
    \textbf{Fei Wu}\textsuperscript{\normalfont 1}
    \\[0.45em]
    \textsuperscript{1}Zhejiang University, Hangzhou, China
    \\
    \textsuperscript{2}National University of Defense Technology,
    Changsha, China
    \\
    \textsuperscript{3}Alibaba Group, Hangzhou, China
    \\
    \textsuperscript{4}Northeastern University, Shenyang, China
    \\[0.45em]
    {\normalfont\normalsize
    \textbf{Emails:}
    \texttt{\{huyinghao,wuyiquan,kunkuang\}@zju.edu.cn}
    }
}

\begin{document}
\maketitle
\renewcommand{\thefootnote}{\fnsymbol{footnote}}
\footnotetext[1]{Corresponding authors.}
\renewcommand{\thefootnote}{\arabic{footnote}}

\begin{abstract}
With the rapid development of LLMs, increasing attention has been paid to whether they can reliably generate complete and legally valid judicial documents. This calls for a benchmark capable of evaluating whether models can accurately apply statutes and resolve outcome-critical issues, including claims, charges, penalties, and monetary obligations.  However, existing benchmarks typically focus on isolated legal tasks or coarse-grained evaluation. We introduce ChJGBench, a benchmark for generating the reasoning and dispositive sections of Chinese judgments across civil, criminal, and administrative cases. Built from 3,462 real-world judgments, ChJGBench provides a case-type-aware diagnostic evaluation framework comprising 13 metric types and adopts a hybrid paradigm. Experiments reveal no uniform advantage for legal-domain models over general-purpose models, with model strengths varying across diagnostic dimensions. Performance is generally stronger on criminal cases, while civil and administrative cases continue to present substantial challenges.

\end{abstract}

\section{Introduction}

In China, written judgments are the primary official records through which courts present factual findings, legal reasoning, and dispositive outcomes. Generating their core adjudicative sections requires models not only to produce fluent legal text, but also to resolve outcome-critical elements such as claims, charges, penalties, monetary obligations, and statutory grounds. Prior work has studied criminal court-view generation from fact descriptions~\cite{ye-etal-2018-interpretable,liu-etal-2024-unleashing,xu-etal-2024-divide}. However, existing evaluations still have limitations in coverage and granularity: studies of court-view generation largely focus on criminal cases, while multi-type benchmarks use smaller datasets and comparatively coarse or heterogeneous evaluation criteria. Consequently, it remains unclear how reliably current LLMs generate judicial reasoning and dispositive outcomes across litigation types under a unified, fine-grained diagnostic framework.

\begin{table}[t]
\centering
\small
\setlength{\tabcolsep}{3pt}
\renewcommand{\arraystretch}{1.15}
\begin{tabular}{l c c c c}
\toprule
\textbf{Benchmark} & \textbf{M-Type} & \textbf{Samp.} & \textbf{Dims.} & \textbf{Eval. Method} \\
\midrule
JuDGE      & \texttimes & 501   & 4  & Det. \\
CaseGen    & \texttimes & 500   & 15  & LLM \\
J1-EVAL    & \texttimes & 162   & 6  & Det. + LLM \\
PLawBench  & \text{N/A}         & 40    & 6  & Rubric + LLM \\
\textbf{ChJGBench} & \checkmark & 3,462 & 13 & Diag. + LLM \\
\bottomrule
\end{tabular}
\caption{Comparison of \textbf{ChJGBench} with prior legal judgment and document generation benchmarks.
\textbf{M-Type}: all three litigation types;
\textbf{Samp.}: distinct source cases;
\textbf{Dims.}: dimensions for document drafting and judgment generation;
\textbf{Det.}: deterministic evaluation, e.g., exact, pattern, or regex matching;
\textbf{Diag.}: diagnostic evaluation with litigation-type-specific criteria.}
\label{tab:chjgbench_comparison}
\end{table}

Existing legal benchmarks~\cite{fei-etal-2024-lawbench,chalkidis-etal-2022-lexglue,NEURIPS2023_89e44582,NEURIPS2024_2cb40fc0} have contributed to the systematic evaluation of legal understanding and reasoning, but they remain insufficient for evaluating judgment generation. As summarized in Table~\ref{tab:chjgbench_comparison}, prior benchmarks are often limited in at least one of the following aspects: litigation-type coverage, evaluation scale, diagnostic granularity, or evaluation methodology. Many benchmarks focus on isolated legal tasks, such as classification, multiple-choice question answering, charge prediction, or legal consultation. Even recent benchmarks on judgment or legal document generation typically include fewer samples and evaluation dimensions, or cover only a subset of litigation types. Moreover, commonly used text-similarity metrics such as BLEU~\cite{papineni-etal-2002-bleu} and ROUGE~\cite{lin-2004-rouge} are inadequate for legal evaluation, as surface-level overlap may overlook substantive legal errors, including incorrect charges, missing payment obligations, or wrong statutory provisions. Unconstrained LLM-as-a-judge evaluation may further introduce subjectivity and instability when assessing long and complex judicial documents.

To address these limitations, we introduce \textbf{ChJGBench}, a comprehensive benchmark for Chinese legal judgment generation. ChJGBench contains 3,462 real-world judgments issued between 2019 and 2024, including 1,650 civil cases, 1,022 criminal cases, and 790 administrative cases. Rather than directly sampling from the empirical distribution of collected judgments, which would overrepresent frequent categories and underrepresent long-tail case types, we adopt a coverage-oriented sampling strategy that balances distributional fidelity with category coverage. This design allows ChJGBench to better reflect both common and less frequent adjudicative scenarios.

\begin{table}[t]
\centering
\small
\setlength{\tabcolsep}{5pt}
\renewcommand{\arraystretch}{1.20}

\definecolor{commonbg}{RGB}{235,245,255}
\definecolor{crimbg}{RGB}{255,242,230}
\definecolor{civbg}{RGB}{240,248,238}
\definecolor{admbg}{RGB}{245,240,255}

\begin{tabularx}{\columnwidth}{
    >{\centering\arraybackslash}p{0.2\columnwidth}
    >{\raggedright\arraybackslash}X
    >{\centering\arraybackslash}p{0.13\columnwidth}
}
\toprule
\textbf{Abbr.} & \textbf{Dimension} & \textbf{Section} \\
\midrule

\rowcolor{commonbg}
\textbf{Law} 
& Statutory Provision Application 
& ~\ref{sec:law} \\

\rowcolor{commonbg}
\textbf{ISS} 
& Disputed Issues Identification 
& ~\ref{sec:iss} \\

\rowcolor{commonbg}
\textbf{JUS} 
& Judicial Justification 
& ~\ref{sec:jus} \\
\midrule

\rowcolor{crimbg}
\textbf{Charge} 
& Criminal Charge Prediction 
& ~\ref{sec:charge} \\

\rowcolor{crimbg}
\textbf{Fine} 
& Monetary Fine
& ~\ref{sec:fine} \\

\rowcolor{crimbg}
\textbf{REC} 
& Recidivism Assessment 
& ~\ref{sec:rec} \\

\rowcolor{crimbg}
\textbf{MIT} 
& Mitigating Circumstances
& ~\ref{sec:mit} \\

\rowcolor{crimbg}
\textbf{Pen\_M} 
& Penalty Duration Prediction 
& ~\ref{sec:pen-m} \\
\midrule

\rowcolor{civbg}
\textbf{SUP$_{\mathrm{Civ}}$} 
& Claim Support 
& ~\ref{sec:sup-civ} \\

\rowcolor{civbg}
\textbf{PAY$_{\mathrm{Civ}}$} 
& Payment Obligation
& ~\ref{sec:pay-civ} \\
\midrule

\rowcolor{admbg}
\textbf{SUP$_{\mathrm{Adm}}$} 
& Claim Support 
& ~\ref{sec:sup-adm} \\

\rowcolor{admbg}
\textbf{PAY$_{\mathrm{Adm}}$} 
& Payment Obligation
& ~\ref{sec:pay-adm} \\

\rowcolor{admbg}
\textbf{DIS} 
& Administrative Disposition 
& ~\ref{sec:dis} \\

\bottomrule
\end{tabularx}

\caption{Case-type-aware evaluation dimensions in ChJGBench, grouped by background color into shared, criminal, civil, and administrative dimensions. Each entry reports the metric abbreviation and points to the section where the metric is defined.}
\label{tab:evaluation_framework}
\end{table}

A key feature of ChJGBench is its case-type-aware diagnostic evaluation framework. Instead of assigning a single generic score to generated judgments, we evaluate them along both shared and litigation-type-specific dimensions. The shared dimensions assess statutory provision application, disputed issue identification, and judicial justification. The case-specific dimensions capture distinctive adjudicative requirements in different litigation types: criminal cases involve charge prediction, monetary fine, recidivism assessment, mitigating circumstance identification, and penalty duration prediction; civil cases involve claim support and payment obligation; and administrative cases involve claim support, administrative disposition, and payment obligation. In total, ChJGBench defines 13 distinct metric types: three shared metrics and ten litigation-specific metrics. These correspond to 19 case-type-specific metric instances in total.

To improve evaluation reliability and scalability, ChJGBench adopts a hybrid metric computation paradigm. LLMs are used in a constrained manner for legal-element extraction and normalization. For ISS, JUS, and DIS, an evaluator LLM additionally performs pairwise semantic tagging under dimension-specific rubrics. The resulting structured elements and semantic tags are then converted into scores using deterministic, dimension-specific scoring procedures. In this way, LLMs serve as controlled information extractors and semantic taggers rather than unconstrained subjective judges. The LLM-based extraction prompts for all evaluation dimensions are provided in Appendix~\ref{app:extraction_prompts}, while the semantic-tagging prompts for ISS, JUS, and DIS are provided in Appendix~\ref{app:semantic_matching_prompts}.

Our contributions are threefold. First, we construct \textbf{ChJGBench}, a large-scale benchmark for Chinese legal judgment generation with 3,462 real-world judgments and a coverage-oriented data distribution across civil, criminal, and administrative cases. Second, we propose a case-type-aware diagnostic evaluation framework with 13 fine-grained dimensions tailored to common legal reasoning and litigation-specific adjudicative requirements. Third, we conduct comprehensive experiments on ChJGBench to analyze model performance across litigation types, legal task families, and diagnostic dimensions, providing empirical insights for future research on reliable legal judgment generation.

\section{Related Work}
\subsection{Legal Benchmarks}

With the adoption of LLMs in the legal domain~\cite{deng-etal-2023-syllogistic,shui-etal-2023-comprehensive,Louis_2024}, there is an increasing demand for benchmarks that better reflect real-world legal practice. Existing efforts have primarily focused on evaluating legal understanding and reasoning through structured tasks. For example, LexGLUE~\cite{chalkidis-etal-2022-lexglue} provides a suite of multi-label classification and multiple-choice tasks derived from European and U.S. legal systems. LegalBench~\cite{NEURIPS2023_89e44582} and LEXam~\cite{ICLR2026_47cbdfce} further extend evaluation within case law systems, covering diverse forms of legal reasoning such as rule application and legal interpretation. In the Chinese legal domain, benchmarks such as LawBench~\cite{fei-etal-2024-lawbench} and LexEval~\cite{NEURIPS2024_2cb40fc0} adopt a multi-task paradigm to assess a broad spectrum of capabilities, including charge prediction, simulated bar exam questions, and legal consultation.

These benchmarks primarily focus on natural language understanding, classification, or question answering tasks within the legal domain. However, real-world judicial practice requires more than solving isolated legal tasks: courts must produce complete decisions that integrate fact finding, legal reasoning, and final dispositions. This mismatch highlights the need for benchmarks that evaluate legal judgment generation in a comprehensive and structured manner.

\subsection{Legal Judgment Document Generation}

Prior work on judicial judgment generation has primarily focused on criminal cases. Early studies explore court-view generation from fact descriptions~\cite{ye-etal-2018-interpretable}, while subsequent work decomposes the task according to legal outcomes such as charges, law articles, and prison terms~\cite{liu-etal-2024-unleashing}, or legal concepts such as recidivism and confession~\cite{xu-etal-2024-divide}. Evaluation in these settings often relies in part on general text-similarity metrics, such as ROUGE-L~\cite{lin-2004-rouge} and BERTScore~\cite{Zhang2020BERTScore}. While useful for measuring textual or semantic similarity, such metrics provide limited information about legally consequential distinctions, such as charge determination and sentencing outcomes. More recent work has moved toward broader and more structured evaluation. SimuCourt~\cite{he-etal-2024-agentscourt} covers 420 criminal, civil, and administrative judgments and combines deterministic, LLM-based, and human evaluation. JuDGE~\cite{su2025judge} provides more fine-grained evaluation of charge, penalty, and legal-reference accuracy, although it focuses on criminal judgments. CaseGen~\cite{li2025casegenbenchmarkmultistagelegal}, J1-EVAL~\cite{jia-etal-2026-ready}, and PLawBench~\cite{shi-etal-2026-plawbench} further employ LLM-based or rubric-based evaluation for broader legal generation and drafting tasks.

These studies substantially expand the scope and granularity of legal generation evaluation, but a unified framework for fine-grained evaluation of judicial reasoning and dispositive outcomes across criminal, civil, and administrative judgments remains relatively underexplored. ChJGBench aims to contribute in this direction by evaluating these three litigation types within a unified framework that combines deterministic matching with LLM-based semantic evaluation.

\section{Task Definition}

Chinese judicial documents include judgments, rulings, mediation statements, decisions, notices, orders, and other instruments. Among them, \textbf{judgments} constitute the most representative and practically important category of judicial decisions, as they require enforceable rulings, rigorous legal reasoning, and a standardized document structure. We therefore focus on judgment generation as the core task. To enable systematic modeling and evaluation, we decompose a Chinese court judgment into six logical components and formally define the task accordingly. Figure~\ref{Judgment-Document} illustrates this structure, and further details are provided in Appendix~\ref{judgmentdetail:appendix}.

\begin{equation}
\begin{split}
     \mathcal{D} = (&\mathit{Header},\ \mathit{Background},\\
     &\mathit{Findings\ of\ Fact},\ \mathit{Reasoning},\\
     &\mathit{Judgment},\ \mathit{Footer}).   
\end{split}
\end{equation}

\begin{figure}[t]
	\centering
	\includegraphics[width=.9\linewidth]{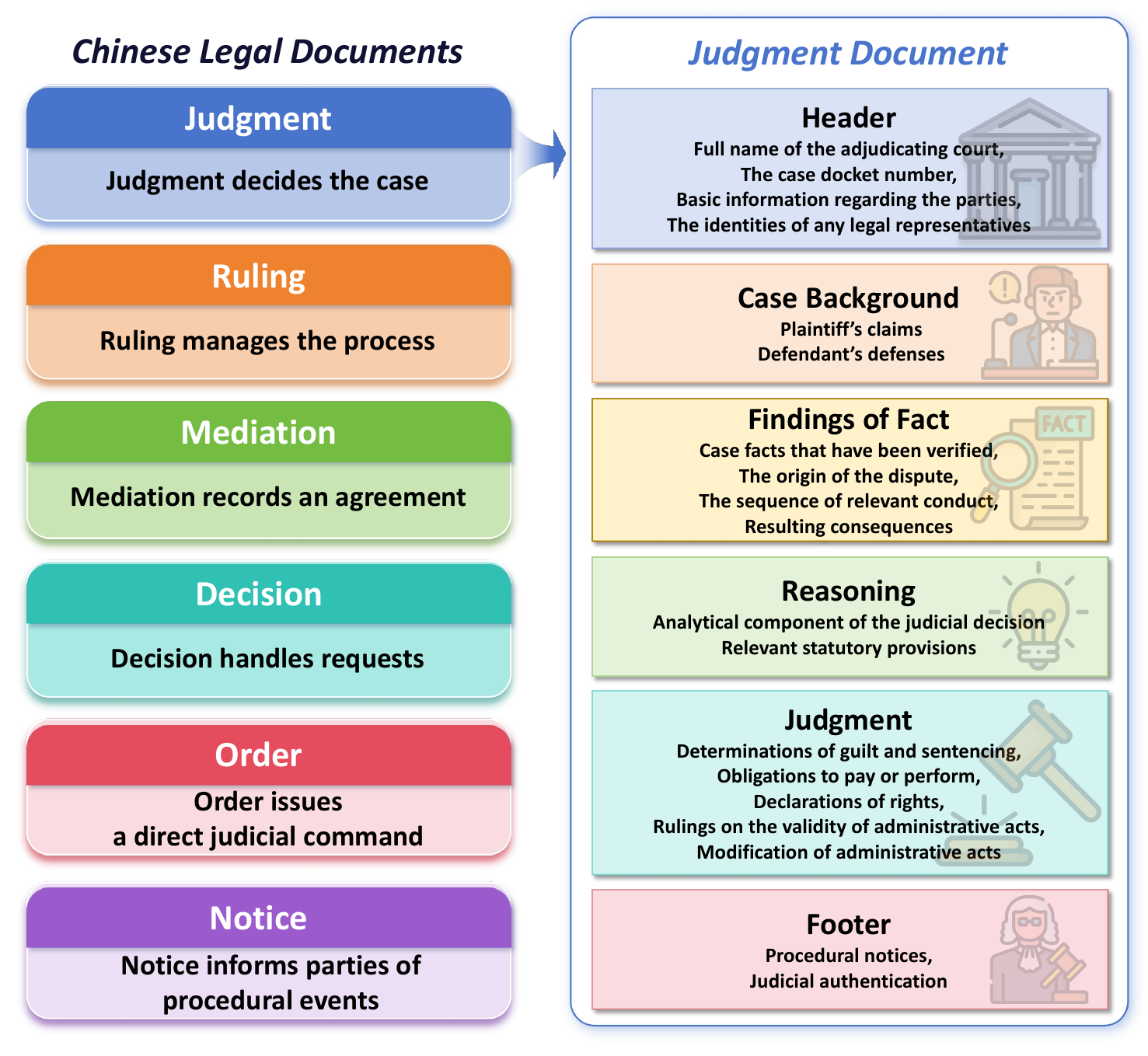} 
	\caption{Structure of a legal judgment document.}
	\label{Judgment-Document}
\end{figure}

We use \textit{Header}, \textit{Background}, and \textit{Findings of Fact} as the input because they provide the procedural and factual context necessary for adjudication. We exclude \textit{Footer} because it is largely formulaic. The generation target comprises \textit{Reasoning} and \textit{Judgment}, as these two sections constitute the core of a judicial decision. The former reflects the legal reasoning process, namely how legal conclusions are derived from established facts, while the latter presents the final adjudicative ruling. Generating these sections requires the model to integrate legal knowledge, perform legal inference, and conform to established conventions of judicial practice, thereby enabling a reliable mapping from factual descriptions to adjudicative outcomes.

In summary, the legal judgment generation task takes \(\mathcal{X}\) as input and produces \(\mathcal{Y}\) as output, where
\begin{equation}
\begin{aligned}
\mathcal{X}
&=(\mathit{Header},\ \mathit{Background},\ \mathit{Findings\ of\ Fact})\\
\mathcal{Y}
&=(\mathit{Reasoning},\ \mathit{Judgment}).
\end{aligned}
\label{eq:judgment_generation_task}
\end{equation}

\section{The ChJGBench Benchmark}
\subsection{Benchmark Construction}

\paragraph{Data Sources}
We construct ChJGBench to evaluate Chinese legal judgment generation across civil, criminal, and administrative cases. All judgments are collected from China Judgments Online\footnote{https://wenshu.court.gov.cn/}, the official national repository of Chinese court documents. Each judgment is parsed into six standard structural components (see Figure~\ref{Judgment-Document}), supporting flexible input construction.

\paragraph{Data Distribution}
Sampling directly from the empirical distribution of collected judgments would overrepresent frequent categories and underrepresent long-tail ones. To balance distributional fidelity with category coverage, we adopt a coverage-oriented sampling strategy. We first stratify the candidate pool by case type \(t\) and issuance year \(y\) from 2019 to 2024. Within each stratum, cases are sampled according to legal category \(c\): cause of action for civil cases, charge for criminal cases, and dispute category for administrative cases.

Let \(P_{t,y}(c)\) denote the empirical category distribution within stratum \((t,y)\). To improve long-tail coverage, we apply a power-temperature transformation:
\begin{equation}
\tilde{P}_{\tau}(c \mid t,y)
=
\frac{P_{t,y}(c)^\tau}
{\sum_k P_{t,y}(k)^\tau},
\quad
\tau \in (0,1].
\end{equation}
When \(\tau < 1\), low-frequency categories receive higher relative sampling probabilities, while dominant categories are down-weighted.

To constrain deviation from the empirical distribution, we interpolate the original and smoothed distributions:
\begin{equation}
\begin{aligned}
Q_{\lambda,\tau}(c\mid t,y)
&=\lambda P_{t,y}(c)\\
&\quad +(1-\lambda)\tilde{P}_{\tau}(c\mid t,y),
\end{aligned}
\end{equation}
where \(\lambda \in [0,1]\) balances empirical fidelity and long-tail coverage. For each case type, instances are sampled according to \(Q_{\lambda,\tau}\) until the target sample size is reached. We set \(\tau=0.5\) and \(\lambda=0.3\) as the sampling hyperparameters; Appendix~\ref{app:sampling_hyperparameters} provides a detailed discussion of their effects.

After structural parsing, we apply quality filters to retain only formal judgment documents, removing duplicates, documents with incomplete structural fields, and cases with failed or ambiguous section segmentation. To ensure sufficient factual and legal content, we require both the \textit{Findings of Fact} and \textit{Court's Reasoning} sections to contain more than 100 Chinese characters. The final benchmark contains 3,462 judgments issued between 2019 and 2024, including 1,650 civil, 1,022 criminal, and 790 administrative cases. Detailed dataset statistics are provided in Appendix~\ref{sec:benchmark_dataset_statistics}.

\subsection{Case-Type-Aware Diagnostic Evaluation Framework}

A central challenge in evaluating legal judgment generation is that a judgment is not simply a fluent piece of legal writing. It is a structured legal decision that must identify disputed issues, apply relevant statutory provisions, provide judicial reasoning, and reach legally valid outcomes. Existing evaluation methods are often insufficient for this setting. Surface-level similarity metrics, such as lexical overlap or embedding similarity, may fail to detect legally decisive errors; for example, a judgment with a correct format but incorrect charges or statutory provisions may still receive a high score. Meanwhile, directly asking an LLM to assign a quality score, even with clear instructions, can introduce subjectivity and instability, especially for long and complex judgments.

As summarized in Table~\ref{tab:evaluation_framework}, the framework does not apply a uniform metric set to all judgments. It organizes the 13 metric types into three shared dimensions and ten litigation-specific dimensions. The shared dimensions capture requirements common to civil, criminal, and administrative cases, while the litigation-specific dimensions reflect the distinctive adjudicative tasks of each litigation type.

Within this structure, the scoring pipeline separates constrained LLM-based processing from deterministic score computation. Under explicit prompts, LLMs extract or normalize predefined legal elements from generated judgments. For ISS, JUS, and DIS, an evaluator LLM performs pairwise semantic tagging under dimension-specific rubrics. The resulting elements and tags are converted into scores using deterministic procedures tailored to each dimension: set-based F$_1$ for Law and Charge; macro-averaged F$_1$ for REC, MIT, and SUP; NVS for Fine and PAY; Pen\_M scoring based on NVS for numerical durations and exact matching for categorical punishments; and soft semantic F$_1$ aggregation for ISS, JUS, and DIS.

By restricting LLMs to controlled extraction, normalization, and rubric-guided semantic tagging, the framework avoids unconstrained holistic scoring. The resulting evaluation is both diagnostic and legally grounded: it assesses whether generated judgments reach correct conclusions on key adjudicative components. Appendix~\ref{app:Evaluation_Components_and_Scoring_Methods} further specifies, for each dimension, the LLM processing components and final scoring method.

\subsection{Shared Dimensions}
\paragraph{Statutory Provision Application (Law).}
\label{sec:law}
This metric evaluates whether the statutory provisions cited in the model-generated "Court's Reasoning" section match the reference provisions. It is applicable to civil, criminal, and administrative cases. Each provision is represented as \texttt{[Statute Name] Article X}. Statute names are standardized using a predefined alias set, and each combination of a statute name and an article number is treated as one item. We compute an F$_1$ score for each case by exact matching and report the mean across cases.

\paragraph{Disputed Issues Identification (ISS).}
\label{sec:iss}
This metric evaluates whether the generated judgment captures the core legal questions under adjudication. These issues vary by case type: they may involve rights and obligations between equal parties in civil cases, criminal prosecution by the state in criminal cases, and the legality of administrative actions in administrative cases. We use an LLM to extract disputed-issue lists from the judgment generated by the model under evaluation and the corresponding real case. Given the source legal text, the model-output extract list, and the ground-truth content list, an evaluator LLM assigns exactly one semantic tag to every possible pair consisting of one ground-truth content item and one model-output extract under the ISS-specific rubric, and the case-level score is computed deterministically from these tags using Equation~\ref{eq:soft_semantic_f1}.

\paragraph{Judicial Justification (JUS).}
\label{sec:jus}
This metric evaluates whether the generated judgment provides the legally material analysis and grounds explaining why the court reaches its disposition, including how adjudicated facts and applicable law support its conclusions on disputed issues, claims, or charges. It measures how well the generated reasoning covers and matches the ground-truth fact-rule-conclusion points, rather than evaluating linguistic fluency or overall textual similarity. We first use an LLM to extract lists of key reasoning points from the judgment generated by the model under evaluation and the corresponding real case. Given the source legal text, the model-output extract list, and the ground-truth content list, an evaluator LLM assigns exactly one semantic tag to every possible pair consisting of one ground-truth content item and one model-output extract according to the JUS-specific rubric, and the case-level score is computed deterministically from these tags using Equation~\ref{eq:soft_semantic_f1}.

\subsection{Criminal Case Dimensions}
\paragraph{Criminal Charge Prediction (Charge).}
\label{sec:charge}
This metric evaluates criminal charge prediction by comparing the charges extracted from the generated judgment with the ground-truth charges for each defendant. We use an LLM to extract the predicted charge labels and compute the F$_1$ score through label matching.

\paragraph{Monetary Fine (Fine).}
\label{sec:fine}
This metric measures the numerical accuracy of monetary fines imposed in criminal cases. For a predicted amount \(a\) and the corresponding reference amount \(b\), we compute Normalized Value Similarity (NVS) as
\begin{equation}
\mathrm{NVS}(a,b)=1-\frac{|a-b|}{\max(a,b)},
\label{eq:NVS}
\end{equation}
where \(a,b\ge 0\) and \(a+b>0\). For cases with multiple defendants, we compute the case-level Fine score by averaging the defendant-level NVS scores over all defendants who receive a fine in the reference judgment. A defendant receives a score of \(0\) if the corresponding fine amount is missing from the prediction. The final Fine score is the mean of the resulting case-level scores.

\paragraph{Recidivism Assessment (REC).}
\label{sec:rec}
This metric evaluates whether the generated judgment correctly identifies each defendant's recidivist status. Recidivism is a legally significant sentencing factor because it may affect sentencing severity and the availability of lenient treatment. We formulate this dimension as a binary classification task and report the macro-averaged F$_1$ score over the recidivist and non-recidivist classes, with each defendant treated as one instance.

\paragraph{Mitigating-Circumstance Identification (MIT).}
\label{sec:mit}
This metric evaluates whether the generated judgment correctly identifies the presence of a mitigating circumstance for each defendant. Such circumstances may include voluntary surrender, confession, remorse, restitution, compensation, or obtaining the victim's forgiveness. We formulate this dimension as a defendant-level binary classification task and report the macro-averaged F$_1$ score over the positive and negative classes.

\paragraph{Penalty Duration Prediction (Pen\_M).}
\label{sec:pen-m}
This metric evaluates whether the generated judgment predicts the correct principal punishment. It covers duration-based punishments, including fixed-term imprisonment, criminal detention, and public surveillance, as well as life imprisonment and the death penalty. Let \(\mathcal{T}\subset\mathbb{R}_{>0}\) denote punishment durations normalized to months, and let \(\mathcal{C}\) contain the categorical labels \texttt{LIFE\_IMPRISONMENT} and \texttt{DEATH\_PENALTY}. For each defendant with a reference punishment, let \(\hat{p}_i\) and \(p_i\) denote the predicted and reference punishments, respectively. The defendant-level score is defined as

\begin{equation}
s_i^{\mathrm{Pen\_M}}=
\begin{cases}
\mathrm{NVS}(\hat{p}_i,p_i), & \hat{p}_i,p_i\in\mathcal{T},\\
1, & \hat{p}_i=p_i\in\mathcal{C},\\
0, & \text{otherwise}.
\end{cases}
\label{eq:penalty_instance_score}
\end{equation}

For each case, the case-level Pen\_M score is the mean of the defendant-level scores for all defendants with an applicable reference punishment. The final Pen\_M score is the mean of the case-level scores over cases containing at least one such defendant.

\subsection{Civil Case Dimensions}

\paragraph{Claim Support (SUP$_{\mathrm{Civ}}$).}
\label{sec:sup-civ}
This metric evaluates whether the generated civil judgment correctly predicts the overall disposition of the plaintiff's claims as upheld, partially upheld, or rejected. It captures the substantive adjudicative outcome of civil litigation. We formulate it as a three-class classification task and report macro-F$_1$ over the three outcome labels.

\paragraph{Payment Obligation (PAY$_{\mathrm{Civ}}$).}
\label{sec:pay-civ}
This metric evaluates whether the generated civil judgment accurately predicts the total amount of payment obligations imposed on each obligor, including compensation, repayment, damages, restitution, and other monetary liabilities. We evaluate obligor-level numerical accuracy using NVS, as defined in Equation~\ref{eq:NVS}, and average the resulting scores first within each case and then across cases. Evaluation is restricted to civil cases containing at least one reference payment obligation.

\subsection{Administrative Case Dimensions}
\paragraph{Claim Support (SUP$_{\mathrm{Adm}}$).}
\label{sec:sup-adm}
This metric evaluates whether the generated administrative judgment correctly predicts the overall disposition of the plaintiff's claims. Although this dimension is also used in civil cases, its legal meaning differs in administrative litigation, where claims typically concern the legality of administrative acts, administrative nonfeasance, or requests for administrative remedies. The disposition is classified as \texttt{upheld}, \texttt{partially upheld}, or \texttt{rejected}, and performance is measured using macro-F$_1$ over the three outcome labels.

\paragraph{Payment Obligation (PAY$_{\mathrm{Adm}}$).}
\label{sec:pay-adm}
This metric evaluates whether the generated administrative judgment accurately predicts the amounts of monetary obligations, such as administrative compensation, restitution, refunds, or other payment-related remedies. We compute PAY$_{\mathrm{Adm}}$ in the same way as PAY$_{\mathrm{Civ}}$.

\paragraph{Administrative Disposition Prediction (DIS).}
\label{sec:dis}
This metric evaluates whether the generated administrative judgment accurately predicts non-monetary administrative dispositions, such as annulment of an administrative act, orders to reissue a decision, or declarations of unlawfulness. We use an LLM to extract lists of atomic core dispositions from the judgment generated by the model under evaluation and the corresponding real case, retaining each disposition's operative action, object, and any explicitly stated scope or condition. Given the source legal text, the model-output extract list, and the ground-truth content list, an evaluator LLM assigns exactly one semantic tag to every possible pair consisting of one ground-truth content item and one model-output extract, and the case-level score is computed deterministically using Equation~\ref{eq:soft_semantic_f1}. Evaluation is restricted to administrative cases with explicit dispositive language in the corresponding real case.

\section{Experiments}

This section presents a comprehensive evaluation of current LLMs on \textbf{ChJGBench}. We compare legal-domain models and strong general-purpose LLMs across civil, criminal, and administrative cases under the same diagnostic evaluation framework. Table~\ref{tab:metrics_combined} summarizes the main results over all key metrics. We then introduce the experimental settings, discuss major performance trends, and analyze task difficulty and cross-model variability across evaluation dimensions.

\begin{table*}[t]
\centering
\footnotesize
\setlength{\tabcolsep}{1.8pt}
\renewcommand{\arraystretch}{1.03}

\begin{tabularx}{\textwidth}{@{}l*{9}{Y}@{}}
\toprule
\multirow{2}{*}{\textbf{Case / Metric}}
& \multicolumn{9}{c}{\textbf{Model}} \\
\cmidrule(lr){2-10}
& \textbf{WisdomI.}
& \textbf{LegalOne}
& \makecell{\textbf{Claude}\\\textbf{Sonnet 4.5}}
& \textbf{DS-R1}
& \makecell{\textbf{Gemini}\\\textbf{3 Pro}}
& \textbf{GPT-5.4}
& \textbf{Grok-4}
& \makecell{\textbf{LLaMA}\\\textbf{405B}}
& \makecell{\textbf{Qwen3}\\\textbf{Max}} \\
\midrule

\multicolumn{10}{l}{\textbf{Criminal}} \\

\quad Law
& \meanstd{0.393}{0.006}
& \secondmeanstd{0.718}{0.007}
& \meanstd{0.677}{0.003}
& \meanstd{0.687}{0.004}
& \bestmeanstd{0.746}{0.002}
& \meanstd{0.642}{0.003}
& \meanstd{0.610}{0.002}
& \meanstd{0.504}{0.006}
& \meanstd{0.643}{0.003} \\

\quad JUS
& \meanstd{0.369}{0.025}
& \meanstd{0.714}{0.002}
& \meanstd{0.734}{0.009}
& \bestmeanstd{0.758}{0.002}
& \meanstd{0.738}{0.001}
& \meanstd{0.702}{0.001}
& \secondmeanstd{0.739}{0.002}
& \meanstd{0.639}{0.003}
& \meanstd{0.734}{0.005} \\

\quad ISS
& \meanstd{0.493}{0.020}
& \meanstd{0.802}{0.003}
& \meanstd{0.812}{0.005}
& \meanstd{0.814}{0.006}
& \secondmeanstd{0.823}{0.000}
& \meanstd{0.786}{0.002}
& \bestmeanstd{0.824}{0.002}
& \meanstd{0.702}{0.010}
& \meanstd{0.797}{0.003} \\

\quad Charge
& \meanstd{0.468}{0.028}
& \meanstd{0.973}{0.000}
& \secondmeanstd{0.974}{0.003}
& \meanstd{0.970}{0.004}
& \secondmeanstd{0.974}{0.003}
& \meanstd{0.973}{0.002}
& \bestmeanstd{0.977}{0.001}
& \meanstd{0.960}{0.004}
& \meanstd{0.972}{0.004} \\

\quad Fine
& \meanstd{0.134}{0.001}
& \meanstd{0.564}{0.001}
& \meanstd{0.577}{0.001}
& \meanstd{0.562}{0.001}
& \bestmeanstd{0.611}{0.001}
& \meanstd{0.574}{0.001}
& \meanstd{0.502}{0.002}
& \meanstd{0.476}{0.002}
& \secondmeanstd{0.594}{0.001} \\

\quad REC
& \meanstd{0.108}{0.000}
& \secondmeanstd{0.736}{0.007}
& \meanstd{0.704}{0.001}
& \meanstd{0.713}{0.005}
& \bestmeanstd{0.753}{0.002}
& \meanstd{0.717}{0.005}
& \meanstd{0.667}{0.007}
& \meanstd{0.496}{0.017}
& \meanstd{0.705}{0.012} \\

\quad MIT
& \meanstd{0.365}{0.038}
& \meanstd{0.967}{0.000}
& \meanstd{0.966}{0.002}
& \meanstd{0.960}{0.002}
& \bestmeanstd{0.974}{0.001}
& \meanstd{0.963}{0.001}
& \secondmeanstd{0.971}{0.003}
& \meanstd{0.882}{0.003}
& \meanstd{0.970}{0.002} \\

\quad Pen\_M
& \meanstd{0.604}{0.003}
& \meanstd{0.731}{0.000}
& \meanstd{0.750}{0.001}
& \meanstd{0.740}{0.000}
& \bestmeanstd{0.786}{0.000}
& \meanstd{0.748}{0.001}
& \meanstd{0.724}{0.001}
& \meanstd{0.602}{0.001}
& \secondmeanstd{0.760}{0.004} \\

\quad \textbf{Macro Avg}
& \meanstd{0.367}{0.007}
& \secondmeanstd{0.776}{0.000}
& \meanstd{0.774}{0.002}
& \secondmeanstd{0.776}{0.001}
& \bestmeanstd{0.801}{0.001}
& \meanstd{0.763}{0.001}
& \meanstd{0.752}{0.001}
& \meanstd{0.658}{0.002}
& \meanstd{0.772}{0.002} \\

\midrule

\multicolumn{10}{l}{\textbf{Civil}} \\

\quad Law
& \meanstd{0.079}{0.013}
& \meanstd{0.424}{0.003}
& \meanstd{0.442}{0.002}
& \meanstd{0.232}{0.006}
& \bestmeanstd{0.595}{0.002}
& \secondmeanstd{0.465}{0.003}
& \meanstd{0.397}{0.006}
& \meanstd{0.173}{0.003}
& \meanstd{0.216}{0.002} \\

\quad JUS
& \meanstd{0.402}{0.043}
& \meanstd{0.590}{0.005}
& \bestmeanstd{0.722}{0.003}
& \meanstd{0.698}{0.001}
& \meanstd{0.657}{0.003}
& \meanstd{0.635}{0.003}
& \secondmeanstd{0.712}{0.003}
& \meanstd{0.511}{0.017}
& \meanstd{0.650}{0.004} \\

\quad ISS
& \meanstd{0.622}{0.029}
& \meanstd{0.741}{0.001}
& \meanstd{0.814}{0.002}
& \secondmeanstd{0.821}{0.002}
& \meanstd{0.812}{0.002}
& \meanstd{0.761}{0.003}
& \meanstd{0.818}{0.005}
& \meanstd{0.745}{0.009}
& \bestmeanstd{0.826}{0.002} \\

\quad SUP$_{\mathrm{Civ}}$
& \meanstd{0.626}{0.029}
& \meanstd{0.786}{0.001}
& \meanstd{0.780}{0.004}
& \meanstd{0.782}{0.004}
& \bestmeanstd{0.828}{0.005}
& \secondmeanstd{0.800}{0.003}
& \meanstd{0.760}{0.001}
& \meanstd{0.685}{0.006}
& \meanstd{0.741}{0.003} \\

\quad PAY$_{\mathrm{Civ}}$
& \meanstd{0.245}{0.004}
& \meanstd{0.744}{0.004}
& \meanstd{0.747}{0.004}
& \secondmeanstd{0.756}{0.004}
& \bestmeanstd{0.758}{0.003}
& \meanstd{0.747}{0.004}
& \meanstd{0.711}{0.004}
& \meanstd{0.617}{0.016}
& \meanstd{0.734}{0.002} \\

\quad \textbf{Macro Avg}
& \meanstd{0.395}{0.023}
& \meanstd{0.657}{0.002}
& \secondmeanstd{0.701}{0.001}
& \meanstd{0.658}{0.001}
& \bestmeanstd{0.730}{0.001}
& \meanstd{0.682}{0.002}
& \meanstd{0.680}{0.003}
& \meanstd{0.546}{0.008}
& \meanstd{0.634}{0.001} \\

\midrule

\multicolumn{10}{l}{\textbf{Administrative}} \\

\quad Law
& \meanstd{0.115}{0.024}
& \meanstd{0.379}{0.004}
& \secondmeanstd{0.563}{0.005}
& \meanstd{0.542}{0.004}
& \bestmeanstd{0.578}{0.005}
& \meanstd{0.516}{0.002}
& \meanstd{0.544}{0.005}
& \meanstd{0.257}{0.009}
& \meanstd{0.429}{0.003} \\

\quad JUS
& \meanstd{0.305}{0.019}
& \meanstd{0.597}{0.007}
& \bestmeanstd{0.735}{0.005}
& \meanstd{0.716}{0.003}
& \meanstd{0.668}{0.003}
& \meanstd{0.638}{0.006}
& \secondmeanstd{0.731}{0.007}
& \meanstd{0.520}{0.007}
& \meanstd{0.689}{0.001} \\

\quad ISS
& \meanstd{0.532}{0.018}
& \meanstd{0.724}{0.002}
& \meanstd{0.798}{0.004}
& \bestmeanstd{0.807}{0.002}
& \meanstd{0.789}{0.002}
& \meanstd{0.745}{0.003}
& \meanstd{0.795}{0.003}
& \meanstd{0.691}{0.002}
& \secondmeanstd{0.801}{0.004} \\

\quad SUP$_{\mathrm{Adm}}$
& \meanstd{0.685}{0.006}
& \meanstd{0.808}{0.008}
& \meanstd{0.825}{0.002}
& \secondmeanstd{0.841}{0.006}
& \bestmeanstd{0.852}{0.003}
& \meanstd{0.826}{0.002}
& \meanstd{0.825}{0.002}
& \meanstd{0.744}{0.001}
& \meanstd{0.829}{0.005} \\

\quad PAY$_{\mathrm{Adm}}$
& \meanstd{0.037}{0.003}
& \meanstd{0.351}{0.005}
& \meanstd{0.368}{0.006}
& \meanstd{0.222}{0.005}
& \bestmeanstd{0.441}{0.001}
& \secondmeanstd{0.429}{0.004}
& \meanstd{0.307}{0.005}
& \meanstd{0.227}{0.004}
& \meanstd{0.311}{0.003} \\

\quad DIS
& \meanstd{0.332}{0.020}
& \meanstd{0.574}{0.008}
& \meanstd{0.765}{0.001}
& \secondmeanstd{0.787}{0.003}
& \meanstd{0.732}{0.004}
& \meanstd{0.701}{0.005}
& \bestmeanstd{0.796}{0.004}
& \meanstd{0.713}{0.003}
& \meanstd{0.685}{0.008} \\

\quad \textbf{Macro Avg}
& \meanstd{0.334}{0.015}
& \meanstd{0.572}{0.003}
& \secondmeanstd{0.676}{0.003}
& \meanstd{0.653}{0.003}
& \bestmeanstd{0.677}{0.003}
& \meanstd{0.642}{0.002}
& \meanstd{0.666}{0.001}
& \meanstd{0.525}{0.001}
& \meanstd{0.624}{0.003} \\

\bottomrule
\end{tabularx}

\caption{Model performance across evaluation dimensions and case types. Values are reported as mean $\pm$ sample standard deviation over three runs. Macro Avg is computed per run as the unweighted mean of the unrounded component scores within each case type. Values are rounded to three decimals. Bold and underlined values indicate the best and second-best means, respectively; means that round to the same three-decimal value share the same rank. Abbreviations: WisdomI. = WisdomInterrogatory; DS-R1 = DeepSeek-R1.}
\label{tab:metrics_combined}
\end{table*}

\subsection{Experimental Settings}

\paragraph{Models.}
The experiments examine two categories of LLMs on legal judgment generation tasks: legal-domain LLMs and mainstream general-purpose LLMs. The two legal-domain LLMs are WisdomInterrogatory~\cite{wu2026wisdominterrogatoryluwenopensourcelegal} and LegalOne~\cite{li2026legalonefamilyfoundationmodels}. The seven mainstream general-purpose LLMs are Claude Sonnet 4.5~\cite{anthropic2025claudesonnet45}, DeepSeek-R1~\cite{Guo_2025}, Gemini 3 Pro Preview~\cite{googledeepmind2026gemini3pro}, GPT-5.4~\cite{openai2026gpt54}, Grok-4~\cite{xai2025grok4}, LLaMA 3.1 405B~\cite{grattafiori2024llama}, and Qwen3-Max~\cite{yang2025qwen3technicalreport}.

\paragraph{Metrics.}
We evaluate model outputs using the case-type-aware metric framework described above. Table~\ref{tab:evaluation_framework} summarizes the evaluation dimensions, and Appendix Table~\ref{tab:Evaluation_components_and_scoring_methods} lists their scoring methods. All scores are reported to three decimal places.

\paragraph{Evaluation Model Validation.}
To assess model outputs in a consistent manner, we use Qwen3-Plus as the evaluation model for information extraction and necessary semantic matching. The extraction accuracy of Qwen3-Plus and its agreement with human annotations in semantic matching are reported in Appendix~\ref{app:extraction_accuracy} and Appendix~\ref{app:semantic_matching}, respectively. Specifically, Qwen3-Plus achieves an average extraction accuracy of 99.41\%, demonstrating reliable and stable structured extraction performance. For the three tasks that require semantic matching, Qwen3-Plus also shows strong agreement with human annotations, with Cohen's $\kappa$ values of 0.7036 for JUS, 0.8203 for ISS, and 0.8909 for DIS.

\paragraph{Experimental Details.}
We use few-shot prompting for all tasks, with the detailed generation prompt provided in Appendix~\ref{app:generation_prompt}. To ensure fairness and comparability, all models use the same prompt template, demonstration examples, input context, and output constraints. For judgment generation in the main benchmark experiments, the decoding temperature is set to 0 whenever temperature control is supported.

\subsection{Main Results}

\paragraph{Specialized legal models do not necessarily outperform large-scale general models.} LegalOne, a recent 8B legal-domain model, performs competitively on criminal tasks. However, in civil and administrative cases, it is often surpassed by general models such as Gemini 3 Pro Preview and Claude Sonnet 4.5. This pattern suggests that training small-scale models on legal-domain data may help incorporate legal knowledge and improve performance on certain tasks, but such models still fall short of mainstream large-scale models on complex legal reasoning and numerical judgment.

\paragraph{Model advantages are task-specific rather than universal.} Although Gemini 3 Pro Preview obtains the highest average scores across criminal, civil, and administrative cases, there remains room for improvement across many evaluation dimensions. DeepSeek-R1 leads on criminal JUS, while Grok-4 shows strong performance on criminal ISS and Charge, as well as administrative DIS. Claude Sonnet 4.5 performs best on civil and administrative JUS, and Qwen3-Max obtains the highest score on civil ISS. This pattern suggests that different models capture different aspects of legal judgment generation. Some models show stronger capabilities in JUS and ISS, whereas others are more effective in predicting concrete legal outcomes.

\paragraph{Criminal cases show relatively stronger performance, whereas civil and administrative cases remain comparatively challenging.} Focusing on Law, JUS, and ISS, the three dimensions shared across all litigation types, we observe a descriptive pattern: models generally achieve higher average scores on criminal cases, followed by administrative cases, while obtaining lower scores on civil cases. One possible explanation is that criminal judgments may be relatively more homogeneous in structure and reasoning patterns, which may make the mapping from established facts to legal conclusions more consistent across cases. In contrast, civil and administrative judgments may involve more heterogeneous legal relationships, claims, procedural questions, and remedies. These differences may partly account for the observed variation in model performance across litigation types.

\subsection{Analysis of Task Difficulty and Cross-Model Variability}
\begin{figure}[t]
\centering
\includegraphics[width=\columnwidth]{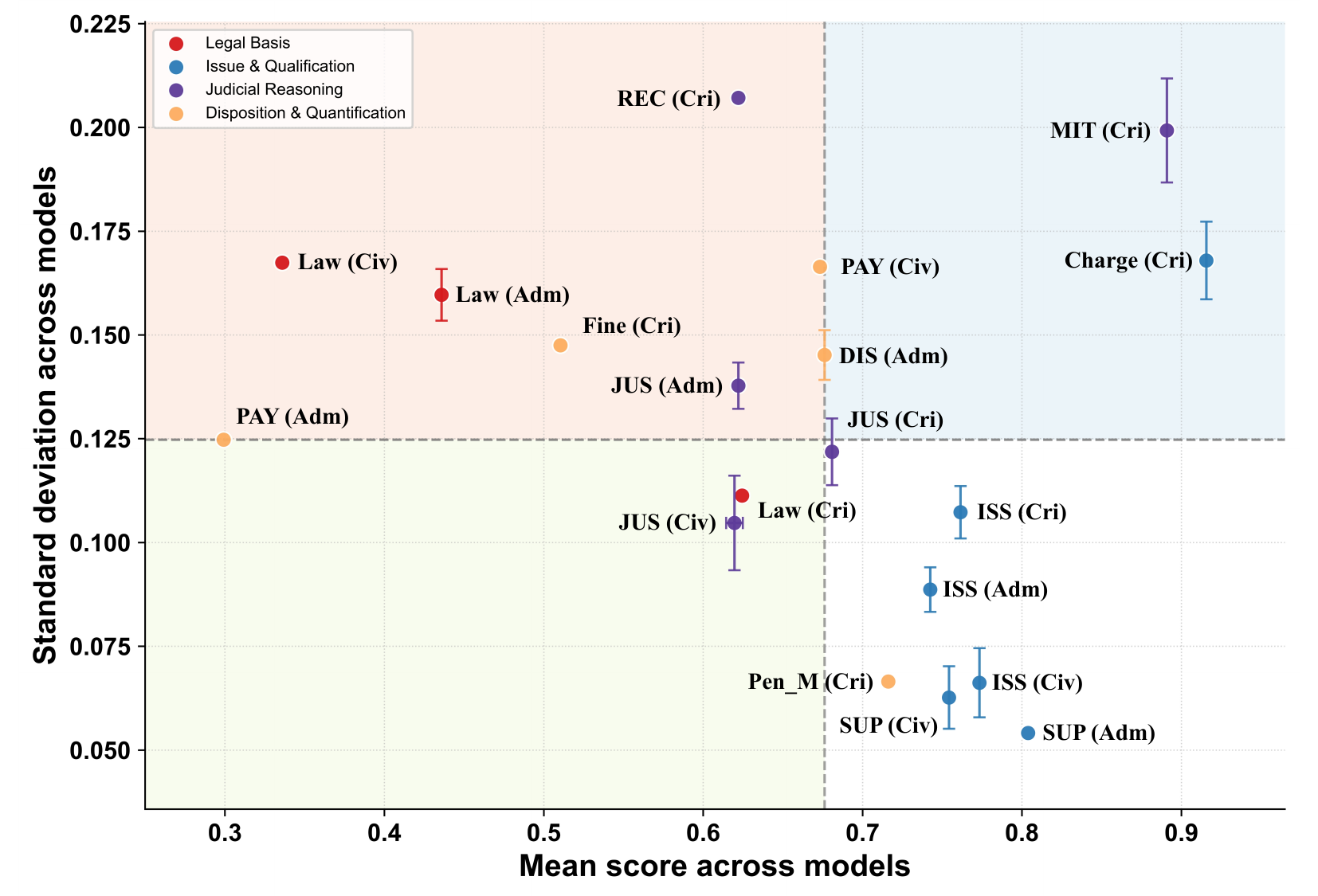}
\caption{
Task difficulty and cross-model variability across evaluation dimensions.
}
\label{fig:task_difficulty_disagreement}
\end{figure}

As shown in Figure~\ref{fig:task_difficulty_disagreement}, each point represents a case-type-specific task. The horizontal coordinate is the unweighted mean of the nine model-level mean scores reported in Table~\ref{tab:metrics_combined}, while the vertical coordinate is their sample standard deviation. The dashed vertical and horizontal lines mark the respective medians of these quantities across the 19 tasks. Point colors indicate the corresponding legal ability families, whose taxonomy is provided in Appendix~\ref{app:LegalAbilityFamilies}. The error bars show how variability across the three runs propagates to the task-level mean and cross-model standard deviation; their calculation is detailed in Appendix~\ref{app:figure2_construction}. For visual clarity, error-bar half-widths below 0.005 are omitted independently for each axis.

The four quadrants reflect different evaluation implications. The left-side quadrants indicate common limitations of current models, as tasks in these regions have relatively low mean scores. From a task-family perspective, \textbf{\textit{Legal Basis}} is the most evident weakness: all Law-related dimensions are located on the left side, showing that statutory grounding remains difficult across case types. This is especially pronounced for Law (Civ) and Law (Adm), suggesting that models struggle to accurately ground generated judgments in domain-specific legal norms. In addition, PAY (Adm), Fine (Cri), and REC (Cri) also appear on the left side, indicating that administrative payment-amount prediction, criminal fine-amount prediction, and recidivism assessment remain challenging tasks.

The upper-right quadrant contains tasks with relatively high mean scores and high cross-model variability. These tasks are not uniformly difficult, as some models can already perform well on them; however, they remain \textbf{relatively discriminative}. In other words, they can serve as useful benchmark indicators. Specifically, metrics such as Charge (Cri) and MIT (Cri) may help distinguish models with stronger legal qualification and adjudicative reasoning capabilities.

In contrast, the lower-right quadrant includes dimensions with high mean scores and low cross-model variability, such as ISS across all case types and SUP in civil and administrative cases. These tasks appear comparatively easier for current models, suggesting that models can generally identify disputed issues and claim support. However, their low cross-model variability makes them \textbf{less effective at differentiating the capabilities} of advanced models.

\section{Conclusions}
In this paper, we introduce \textbf{ChJGBench}, a comprehensive and diagnostically grounded benchmark for Chinese legal judgment generation. ChJGBench is built from 3,462 diverse real-world judgments across civil, criminal, and administrative cases and adopts a coverage-oriented sampling strategy to capture both common and long-tail adjudicative scenarios. It further provides a case-type-aware diagnostic evaluation framework with 13 substantive dimensions and a hybrid rule-and-LLM evaluation paradigm, enabling fine-grained, interpretable, and reliable assessment of both shared legal reasoning abilities and litigation-specific adjudicative outcomes. Our experiments provide several key insights into the current capabilities of LLMs for Chinese legal judgment generation. First, specialized legal models do not consistently outperform large-scale general-purpose models, indicating that legal-domain training alone is not sufficient for complex judgment generation. Second, model strengths are task-specific rather than universal, with different models excelling on different diagnostic metrics. Third, performance varies substantially across litigation types: models achieve higher overall scores on criminal cases but perform less strongly on civil and administrative cases. These results highlight the need to improve models' ability to address the diverse legal relationships, claims, remedies, and statutory requirements involved in civil and administrative adjudication.

\section*{Limitations}

ChJGBench has several limitations. First, because it is constructed from publicly available court judgments, it may not fully represent cases that are unpublished or otherwise unavailable, potentially introducing selection bias. Second, the current version of ChJGBench constitutes a static snapshot, whereas statutes, judicial interpretations, and adjudicatory standards may evolve over time. To partially address this limitation, the reusable construction and stratified sampling pipeline proposed in this paper can be rerun periodically to incorporate new judgments, emerging case categories, and relevant legal developments, thereby enabling targeted and versioned updates. Third, the use of publicly available judgments creates a potential risk of data contamination. Although we remove duplicate documents and apply structural and quality controls, we cannot fully determine whether benchmark cases, closely related cases, or derivative materials were included in the training data of the evaluated models, particularly for closed-source models.

\section*{Ethical Statement}

ChJGBench is constructed from publicly available court judgments collected from China Judgments Online, the official national repository of Chinese court documents. Although the source documents are publicly accessible, judicial records may contain sensitive or potentially identifying information. As part of the release process, we remove or mask personally identifiable and reverse-searchable information and conduct an additional privacy review prior to any public release to reduce privacy and re-identification risks.

ChJGBench is intended solely for research on the capabilities and limitations of language models in legal judgment generation. It must not be used to support or justify real-world judicial decision-making without rigorous human oversight, legal validation, institutional accountability, and compliance with applicable laws and regulations. Benchmark performance should not be interpreted as evidence of deployment readiness or as a substitute for professional legal judgment. We also acknowledge that judicial data may reflect historical, institutional, or social biases, which models trained on or evaluated against such data may reproduce.

\section*{Acknowledgements}
This work was supported by the "Pioneer" and "Leading Goose" R\&D Program of Zhejiang (2025C02037), Research Project of the Philosophy and Social Sciences Planning Research Platform (Laboratory) of Zhejiang Province, and Key R\&D Program of Hangzhou (2025SZDA0254). It was also supported by the National Natural Science Foundation of China (62106039) and Alibaba Research Intern Program. Any opinions expressed in this paper are those of the authors and do not necessarily reflect the views of the funding organizations.

We declare that AI-based writing tools were used only for limited language polishing and grammatical refinement in preparing this paper. The authors have carefully verified all AI-assisted text and take full responsibility for the content of the paper.

\bibliography{custom}
\appendix

\section{Judgment Details}
\label{judgmentdetail:appendix}

\begin{itemize}
    \item \textit{Header}: Contains basic procedural information, including the court name, case number, and the identities of the parties and their representatives.
    \item \textit{Background}: Summarizes the claims, defenses, or charges raised by the parties at the outset of the litigation.
    \item \textit{Findings of Fact}: Presents the facts established through evidence, along with a brief assessment of the supporting evidence.
    \item \textit{Reasoning}: Provides the court's legal analysis by applying relevant statutes to the facts and evaluating the parties' arguments.
    \item \textit{Judgment}: States the court's final and binding ruling in a clear and enforceable format.
    \item \textit{Footer}: Includes procedural notices, such as appeal rights, judicial signatures, the date, and the court seal.
\end{itemize}

\begin{table*}[t]
\centering
\small
\setlength{\tabcolsep}{8pt}
\renewcommand{\arraystretch}{1.25}

\begingroup
\renewcommand{\tabularxcolumn}[1]{m{#1}}
\begin{tabularx}{\textwidth}{
@{}
>{\raggedright\arraybackslash}m{4.0cm}
>{\centering\arraybackslash}m{1.5cm}
>{\raggedright\arraybackslash}X
@{}
}
\toprule
\textbf{Hyperparameter}
& \textbf{Setting}
& \textbf{Function} \\
\midrule

Power-temperature coefficient \(\tau\)
& 0.5
& Controls the degree of smoothing applied to the empirical category distribution. \\

Interpolation coefficient \(\lambda\)
& 0.3
& Controls the balance between the empirical distribution and the temperature-smoothed distribution. \\

\bottomrule
\end{tabularx}
\endgroup

\caption{Hyperparameter settings used in the coverage-oriented sampling strategy. Setting \(\tau=0.5\) applies a square-root transformation to the empirical category distribution, reducing the relative dominance of high-frequency categories and increasing the relative sampling probabilities of low-frequency categories. With \(\lambda=0.3\), the final sampling distribution combines 30\% of the empirical distribution with 70\% of the temperature-smoothed distribution, thereby emphasizing long-tail coverage while retaining the overall structure of the empirical distribution.}
\label{tab:sampling_hyperparameters}
\end{table*}

\section{Hyperparameters for Benchmark Data Distribution}
\label{app:sampling_hyperparameters}

\paragraph{Hyperparameter Settings.}
The hyperparameters used in our coverage-oriented sampling strategy for constructing the benchmark data distribution are summarized in Table~\ref{tab:sampling_hyperparameters}.

\paragraph{Trade-off Interpretation.}
More generally, the formulation provides a tunable trade-off between preserving the empirical category distribution and shifting the sampling distribution toward uniformity across legal categories. When either \(\tau=1\) or \(\lambda=1\), the resulting sampling distribution is identical to the empirical distribution. Conversely, when \(\lambda=0\) and \(\tau\) approaches zero from above, the sampling distribution approaches a uniform distribution over the observed categories within each litigation-type-and-year stratum.

\section{Evaluation Components and Scoring Methods}
\label{app:Evaluation_Components_and_Scoring_Methods}

Table~\ref{tab:Evaluation_components_and_scoring_methods} provides a compact summary of all evaluation dimensions, indicating whether each dimension involves (1) LLM-based extraction or normalization of structured legal elements, (2) LLM-based semantic matching between the generated and reference judgments, and (3) the final scoring method.

\begin{table*}[t]
\centering
\small
\setlength{\tabcolsep}{4pt}
\renewcommand{\arraystretch}{1.15}

\begin{tabularx}{\textwidth}{
    >{\centering\arraybackslash}p{0.105\textwidth}
    >{\centering\arraybackslash}p{0.145\textwidth}
    >{\centering\arraybackslash}p{0.195\textwidth}
    >{\centering\arraybackslash}X
    >{\centering\arraybackslash}p{0.185\textwidth}
}
\toprule

\textbf{Dimension}
&
\textbf{\shortstack{LLM Extraction}}
&
\textbf{\shortstack{LLM Semantic Matching}}
&
\textbf{Final Scoring Method}
&
\textbf{\shortstack{Main-text Section}}
\\

\midrule

\textbf{Law}
& \checkmark
& $\times$
& Case-level set F$_1$
& Sec.~\ref{sec:law}
\\

\textbf{ISS}
& \checkmark
& \checkmark
& Soft semantic F$_1$
& Sec.~\ref{sec:iss}
\\

\textbf{JUS}
& \checkmark
& \checkmark
& Soft semantic F$_1$
& Sec.~\ref{sec:jus}
\\

\midrule

\textbf{Charge}
& \checkmark
& $\times$
& Defendant-level set F$_1$
& Sec.~\ref{sec:charge}
\\

\textbf{Fine}
& \checkmark
& $\times$
& NVS
& Sec.~\ref{sec:fine}
\\

\textbf{REC}
& \checkmark
& $\times$
& Binary macro-F$_1$
& Sec.~\ref{sec:rec}
\\

\textbf{MIT}
& \checkmark
& $\times$
& Binary macro-F$_1$
& Sec.~\ref{sec:mit}
\\

\textbf{Pen\_M}
& \checkmark
& $\times$
& NVS \& exact match
& Sec.~\ref{sec:pen-m}
\\

\midrule

\textbf{SUP$_{\mathrm{Civ}}$}
& \checkmark
& $\times$
& Three-class macro-F$_1$
& Sec.~\ref{sec:sup-civ}
\\

\textbf{PAY$_{\mathrm{Civ}}$}
& \checkmark
& $\times$
& NVS
& Sec.~\ref{sec:pay-civ}
\\

\midrule

\textbf{SUP$_{\mathrm{Adm}}$}
& \checkmark
& $\times$
& Three-class macro-F$_1$
& Sec.~\ref{sec:sup-adm}
\\

\textbf{PAY$_{\mathrm{Adm}}$}
& \checkmark
& $\times$
& NVS
& Sec.~\ref{sec:pay-adm}
\\

\textbf{DIS}
& \checkmark
& \checkmark
& Soft semantic F$_1$
& Sec.~\ref{sec:dis}
\\

\bottomrule
\end{tabularx}

\caption{Evaluation components and scoring methods for each ChJGBench evaluation dimension. LLM extraction denotes the structured extraction or normalization of predefined legal elements. LLM semantic matching denotes constrained element-level comparison between the generated and reference judgments.}
\label{tab:Evaluation_components_and_scoring_methods}
\end{table*}

\section{Validation of Qwen3-Plus as an Evaluation Tool}
Since Qwen3-Plus is used to assist the evaluation of ChJGBench, we conduct two validation experiments to examine its reliability under different evaluation settings. The first experiment evaluates the extraction accuracy of Qwen3-Plus on objectively verifiable legal elements, where model outputs can be compared against manually checked references. The second experiment examines its ability to perform legal semantic matching, focusing on evaluation dimensions that require meaning-level comparison rather than exact string matching.

\subsection{Validation of Qwen3-Plus for Extraction Accuracy}
\label{app:extraction_accuracy}

In ChJGBench, Qwen3-Plus is used to extract or normalize predefined legal elements from generated judgments. To provide a supporting empirical check of this process, we conduct an extraction accuracy validation experiment on a sample of manually verifiable extraction instances.

\paragraph{Data Sampling.}
We randomly sample 1,000 cases from the full benchmark evaluation set to validate the extraction capability of Qwen3-Plus. Each sampled case can be used to construct multiple extraction tasks. Ultimately, we construct 5,000 evaluation instances in the format of {\textit{text segment}, \textit{extraction requirement}, \textit{ground truth}}.
We then provide these 5,000 instances to Qwen3-Plus and require the model to extract the answer according to the given extraction requirement. The model is explicitly instructed not to rewrite or paraphrase the content, but to directly extract the corresponding answer from the original text.

\paragraph{Evaluation Metric.}

We use instance-level accuracy as the validation metric. An extraction is counted as correct when it satisfies the corresponding extraction requirement and matches the manually checked ground truth. Accuracy is defined as:

\begin{equation}
    \text{Acc}
    =
    \frac{1}{N}
    \sum_{i=1}^{N}
    \mathbb{I}
    \left(
    \hat{y}^{(i)} = y^{(i)}
    \right),
    \label{eq:extraction_acc}
\end{equation}
where $\hat{y}^{(i)}$ denotes the result extracted by Qwen3-Plus, $y^{(i)}$ denotes the ground-truth result, and $N$ denotes the number of evaluated instances.

For fields that require normalization, such as monetary amounts, penalty duration, and statutory provisions, we first normalize the extracted results and the ground-truth results into the same format before comparison. The final extraction accuracy reflects the proportion of instances in which Qwen3-Plus produces an output exactly consistent with the ground truth.

\paragraph{Conclusion.}

We repeat the extraction experiment three times to examine the stability of the extraction process. Across the 5,000 extraction instances, Qwen3-Plus correctly extracts 4,978, 4,961, and 4,972 instances in the three runs, corresponding to accuracies of 99.56\%, 99.22\%, and 99.44\%, respectively. The average extraction accuracy is 99.41\%, providing supporting evidence that Qwen3-Plus performs reliably on the sampled extraction instances.

\subsection{Dimension-Specific Validation of Qwen3-Plus for Semantic Matching}
\label{app:semantic_matching}

This validation experiment assesses whether Qwen3-Plus can reliably assign semantic-consistency labels for ISS, JUS, and DIS. We conduct separate agreement analyses for the three dimensions by comparing the labels assigned by Qwen3-Plus with those provided by human annotators.

\paragraph{Data Sampling.}
For each of ISS, JUS, and DIS, we randomly sample 50 validation instances, yielding 150 instances in total. Each instance consists of a triplet containing the \textit{source legal text}, a \textit{model-output extract}, and the corresponding \textit{ground-truth content}. The model-output extract is extracted by Qwen3-Plus from a judgment generated by the model under evaluation, while the ground-truth content is extracted by Qwen3-Plus from the corresponding real case. Given this triplet, Qwen3-Plus and human annotators independently assess the legal-semantic consistency between the model-output extract and the ground-truth content, using the source legal text as case context.

\paragraph{Agreement Labels.}
We ask Qwen3-Plus and three legal professionals to independently annotate the constructed validation instances. For each validation triplet, the annotators use the source legal text as case context and assess the legal-semantic consistency between the model-output extract and the ground-truth content by assigning one of the following three labels:

\begin{itemize}
    \item \textit{\textbf{Equivalent}}: The model-output extract and the ground-truth content are legally equivalent despite possible differences in wording, expression order, or level of detail. They describe the same disputed issue, an equivalent key reasoning point, or the same principal administrative disposition.

    \item \textit{\textbf{Partially Equivalent}}: The model-output extract and the ground-truth content partially overlap in legal meaning, but the model-output extract omits some salient elements, contains minor deviations, or expresses only part of the relevant legal information.

    \item \textit{\textbf{Not Equivalent}}: The model-output extract and the ground-truth content concern different disputed issues, reasoning points, or administrative dispositions and therefore cannot be regarded as semantically consistent in legal meaning.
\end{itemize}

For each validation triplet, the majority label assigned by the three legal professionals is treated as the human annotation and is compared with the label produced by Qwen3-Plus.

\paragraph{Evaluation Metrics.}

We report Cohen's $\kappa$ and Krippendorff's $\alpha$ separately for ISS, JUS, and DIS. Both metrics are computed between the label sequence produced by Qwen3-Plus and the corresponding sequence of human majority labels for each dimension.

\begin{table}[t]
\centering
\small
\setlength{\tabcolsep}{6pt}
\begin{tabular}{ccc}
\toprule
\textbf{Dimension} & \textbf{Cohen's $\kappa$} & \textbf{Krippendorff's $\alpha$} \\
\midrule
ISS & 0.8203 & 0.8216 \\
JUS & 0.7036 & 0.7047 \\
DIS & 0.8909 & 0.8919 \\
\bottomrule
\end{tabular}
\caption{Dimension-specific agreement between Qwen3-Plus and the human majority annotations.}
\label{tab:semantic_matching_validation}
\end{table}

\paragraph{Results and Error Analysis.}
As shown in Table~\ref{tab:semantic_matching_validation}, Qwen3-Plus achieves consistently high agreement with the human majority annotations across the three semantic matching dimensions. The highest agreement is observed for DIS, with a Cohen's $\kappa$ of 0.8909 and a Krippendorff's $\alpha$ of 0.8919. For ISS, the corresponding scores are 0.8203 and 0.8216. JUS is more challenging, but Qwen3-Plus still obtains a Cohen's $\kappa$ of 0.7036 and a Krippendorff's $\alpha$ of 0.7047.

We manually examine all 17 validation instances in which the Qwen3-Plus label differs from the human majority label. The main patterns of disagreement are as follows:

\begin{enumerate}
    \item \textbf{Alternative but legally plausible reasoning in JUS.}
    In some validation triplets, the model-output extract and the ground-truth content follow different but legally plausible reasoning paths given the source legal text. Although these reasoning points may support similar legal conclusions, determining whether they should be classified as fully equivalent, partially equivalent, or not equivalent can be inherently ambiguous. This type of disagreement helps explain the lower agreement observed for JUS compared with ISS and DIS.

    \item \textbf{Differences in disposition granularity in DIS.}
    A small number of disagreements arise when the model-output extract and the ground-truth content convey the same principal administrative disposition but differ in their level of granularity. For example, one may include additional orders, conditions, or implementation details, while the other captures only the principal disposition. Such differences can result in borderline judgments between \textit{Equivalent} and \textit{Partially Equivalent}.
\end{enumerate}

\paragraph{Conclusion.}
The validation results show that Qwen3-Plus achieves substantial agreement with human majority annotations across ISS, JUS, and DIS, with particularly high agreement for ISS and DIS. The remaining disagreements are limited and largely attributable to interpretable edge cases, particularly alternative legal reasoning paths in JUS. These results suggest that Qwen3-Plus can serve as a practical substitute for human semantic matching for these three dimensions under our evaluation setting, supporting scalable and consistent evaluation in ChJGBench.

\section{Legal Ability Family Taxonomy}
\label{app:LegalAbilityFamilies}

To support a more interpretable analysis beyond individual metrics, we group the 13 evaluation metrics into four legal ability families, as summarized in Table~\ref{tab:task_family}. This grouping allows us to examine whether model weaknesses are confined to individual tasks or extend across broader legal capabilities.  The family assignments determine the point colors in Figure~\ref{fig:task_difficulty_disagreement}, allowing task difficulty and cross-model variability to be interpreted at the capability level. The statistical construction of the figure is described in Appendix~\ref{app:figure2_construction}.

\begin{table}[h]
\centering
\small
\setlength{\tabcolsep}{8pt}
\renewcommand{\arraystretch}{1.25}

\begin{tabularx}{0.88\columnwidth}{
    >{\centering\arraybackslash}p{0.3\columnwidth}
    >{\centering\arraybackslash}X
}
\toprule
\textbf{Family} & \textbf{Metrics} \\
\midrule

Legal Basis
& Law \\
\midrule

\makecell[c]{Issue \& \\ Qualification}
& \makecell[c]{Charge, ISS \\ SUP$_{\mathrm{Adm}}$, SUP$_{\mathrm{Civ}}$} \\
\midrule

\makecell[c]{Judicial \\ Reasoning}
& \makecell[c]{JUS, REC \\ MIT} \\
\midrule

\makecell[c]{Disposition \& \\ Quantification}
& \makecell[c]{Fine, Pen\_M, DIS \\ PAY$_{\mathrm{Adm}}$, PAY$_{\mathrm{Civ}}$} \\

\bottomrule
\end{tabularx}

\caption{Legal ability family taxonomy used for task-family analysis.}
\label{tab:task_family}
\end{table}

\section{Task Difficulty and Cross-Model Variability}
\label{app:figure2_construction}

This appendix describes the statistical construction of Figure~\ref{fig:task_difficulty_disagreement}. Each task \(t\) corresponds to a particular evaluation metric for criminal, civil, or administrative cases and is assigned to a legal ability family according to Appendix~\ref{app:LegalAbilityFamilies}.
Let \(\mu_{tm}\) and \(\sigma_{tm}\) denote the mean and sample standard deviation over three runs, respectively, for model \(m\) on task \(t\), as reported in Table~\ref{tab:metrics_combined}. With \(M=9\), the horizontal coordinate is
\begin{equation}
x_t=\frac{1}{M}\sum_{m=1}^{M}\mu_{tm}.
\end{equation}
The vertical coordinate is
\begin{equation}
y_t=\sqrt{\frac{1}{M-1}\sum_{m=1}^{M}(\mu_{tm}-x_t)^2}.
\end{equation}
The dashed vertical and horizontal lines mark the medians of \(x_t\) and \(y_t\), respectively, across the 19 tasks. Assuming independent run-to-run variation across models, the horizontal error-bar half-width is
\begin{equation}
u_{x,t}=\frac{1}{M}\sqrt{\sum_{m=1}^{M}\sigma_{tm}^2}.
\end{equation}
The vertical error-bar half-width is
\begin{equation}
u_{y,t}=\sqrt{\sum_{m=1}^{M}
\left(
\frac{(\mu_{tm}-x_t)\sigma_{tm}}
{(M-1)y_t}
\right)^2}.
\end{equation}
The error bars represent propagated run-to-run variability rather than confidence intervals. Half-widths below 0.005 are omitted independently on each axis.

\section{Benchmark Dataset Statistics}
\label{sec:benchmark_dataset_statistics}

Table~\ref{tab:benchmark_basic_stats} summarizes the basic statistics of the benchmark dataset. The dataset contains 3,462 Chinese legal judgment samples, with an average prompt length of 2,562.95 characters. Each sample includes a reference output comprising the \textit{Reasoning} and \textit{Judgment} sections. The complete reference output, \textit{Reasoning} section, and \textit{Judgment} section have average lengths of 715.48, 595.54, and 119.94 characters, respectively. In addition, each sample cites 4.68 legal provisions on average, indicating that the benchmark requires models not only to generate final judgments but also to ground their predictions in relevant statutory provisions.

\begin{table}[h]
\centering
\small
\setlength{\tabcolsep}{5pt}
\renewcommand{\arraystretch}{1.20}

\definecolor{statbg}{RGB}{235,245,255}

\begin{tabularx}{0.78\columnwidth}{
    >{\raggedright\arraybackslash}X
    >{\raggedleft\arraybackslash}p{0.32\columnwidth}
}
\toprule
\textbf{Statistic (abbr.)} & \textbf{Value} \\
\midrule
\# Samples & 3,462 \\
Lang. & Chinese \\
Prompt len. & 2,562.95 characters \\
Reason len. & 595.54 characters \\
Judgment len. & 119.94 characters \\
Cited provisions & 4.68 \\
\bottomrule
\end{tabularx}

\caption{
Basic statistics of the benchmark dataset. Abbreviations: average prompt length (Prompt len.), average length of the \textit{Reasoning} section (Reason len.), average length of the \textit{Judgment} section (Judgment len.), and average number of cited legal provisions (Cited provisions).
}
\label{tab:benchmark_basic_stats}
\end{table}

\section{Generation Prompt}
\label{app:generation_prompt}

Under the same experimental conditions, all models receive the same system instruction, three demonstrations in the same order, target-input format, and output constraints. The generation input contains only the \textit{Header}, \textit{Background}, and \textit{Findings of Fact}, while the model is required to generate only the \textit{Reasoning} and \textit{Judgment}.

\begin{tcblisting}{
listing only,
breakable,
colback=black!2,
colframe=black!45,
boxrule=0.4pt,
left=1mm,right=1mm,top=1mm,bottom=1mm,
title={Generation Prompt (3-shot Format)},
fonttitle=\bfseries\small,
listing options={style=promptstyle}
}
### System Message

You are completing a research benchmark task for Chinese legal judgment generation. Given the Header, Background, and Findings of Fact from a Chinese judgment, generate only its Reasoning and Judgment sections in Chinese.

The Reasoning is the public-facing judicial analysis that would appear in the written judgment. It should explain how the court legally characterizes the established facts, addresses legally material claims, defenses, charges, sentencing circumstances, or disputed issues when applicable, and applies relevant law to support the disposition. It should present the court's stated legal reasoning rather than any private or hidden reasoning process. Where appropriate, cite the relevant statutory provisions and conclude by leading naturally to the operative ruling.

The Judgment states the court's final operative disposition. It should clearly specify the legal consequences for the relevant parties, such as conviction and sentencing, payment or performance obligations, disposition of civil claims, or the validity, annulment, modification, or required performance of administrative acts. When multiple parties, claims, charges, or obligations are involved, state their respective outcomes separately and unambiguously.

### Task and Requirements

1. Analyze the legally material facts and issues and explain how the applicable law supports the disposition.
2. Address material claims, defenses, charges, sentencing issues, payment obligations, or administrative remedies only when applicable and supported by the input.
3. Do not introduce or alter parties, facts, evidence, dates, amounts, claims, or procedural events. Preserve anonymized names and identifiers exactly as provided.
4. Cite only relevant statutory provisions, using their official titles and article numbers. Do not fabricate uncertain citations.
5. Address multiple parties, defendants, claims, charges, or obligations separately and unambiguously.
6. Write in formal and coherent Chinese judicial style.
7. Output only the Reasoning and Judgment sections. Do not add task explanations, evaluation comments, confidence scores, or Markdown fences.

### Demonstration 1

#### Input

Header:
{demo_1_header}

Background:
{demo_1_background}

Findings of Fact:
{demo_1_findings_of_fact}

#### Output

Reasoning:
{demo_1_reasoning}

Judgment:
{demo_1_judgment}

### Demonstration 2

#### Input

Header:
{demo_2_header}

Background:
{demo_2_background}

Findings of Fact:
{demo_2_findings_of_fact}

#### Output

Reasoning:
{demo_2_reasoning}

Judgment:
{demo_2_judgment}

### Demonstration 3

#### Input

Header:
{demo_3_header}

Background:
{demo_3_background}

Findings of Fact:
{demo_3_findings_of_fact}

#### Output

Reasoning:
{demo_3_reasoning}

Judgment:
{demo_3_judgment}

### Target Case

#### Input

Header:
{target_header}

Background:
{target_background}

Findings of Fact:
{target_findings_of_fact}

#### Required Output

Reasoning:

Judgment:

\end{tcblisting}

\section{Dimension-Specific Extraction and Classification: Prompts and Scoring Procedures}
\label{app:extraction_prompts}

We provide the dimension-specific extraction and classification prompts for all 13 evaluation dimensions, together with the corresponding deterministic scoring procedures. The semantic matching and scoring procedures for ISS, JUS, and DIS are presented separately in Appendix~\ref{app:semantic_matching_prompts}. These prompts are used by Qwen3-Plus for controlled information extraction and classification. Braced expressions such as \texttt{\{text\}} denote runtime placeholders.

\subsection{Shared Dimensions}

\begin{tcblisting}{
  listing only,
  breakable,
  colback=black!2,
  colframe=black!45,
  boxrule=0.4pt,
  left=1mm,right=1mm,top=1mm,bottom=1mm,
  title={Law: Statutory Provision Application},
  fonttitle=\bfseries\small,
  listing options={style=promptstyle}
}
### Task and Requirements
I will provide the Court's Reasoning section of a model-generated judgment. Extract every statutory provision explicitly cited in the input text.

Requirements:
1. Extract only provisions explicitly present in the text. Do not infer or add uncited provisions.
2. Normalize each provision to the form "[Statute Name] Article X", while retaining the official statute name in the source language.
3. When one citation contains multiple article numbers, return each combination of the statute name and an article number separately.
4. Return each unique provision once, in the order in which it first appears.
5. If no statutory provision is explicitly cited, return an empty list [].
6. Return only the required list without any explanation.

### Output Format
Return a strict List[str] in the following form:
["[Statute Name] Article 234", "[Statute Name] Article 67"]

### Input
{text}

Extract the results according to the requirements above:
\end{tcblisting}

The extracted provisions are matched against the provisions in the ground truth. Two items are considered a match when their article numbers are identical and their statute names are identical or belong to the same predefined alias group. Duplicate items are counted only once.

For each case \(c\), let \(\mathrm{TP}_c\) denote the number of matched provisions, \(\mathrm{FP}_c\) the number of provisions appearing only in the extracted result, and \(\mathrm{FN}_c\) the number appearing only in the ground truth. The case-level Law score is
\begin{equation}
F_{1,c}^{\mathrm{Law}}
=
\frac{2\mathrm{TP}_c}
{2\mathrm{TP}_c+\mathrm{FP}_c+\mathrm{FN}_c}.
\label{eq:law_case_score}
\end{equation}
If the ground truth contains at least one provision but the extracted result for the model-generated judgment is empty, the case-level score is \(0\).

Let \(K\) denote the number of evaluated cases. The final Law score is the mean of the case-level scores:
\begin{equation}
\mathrm{Law}
=
\frac{1}{K}
\sum_{c=1}^{K}F_{1,c}^{\mathrm{Law}}.
\label{eq:law_score}
\end{equation}

\begin{tcblisting}{
  listing only,
  breakable,
  colback=black!2,
  colframe=black!45,
  boxrule=0.4pt,
  left=1mm,right=1mm,top=1mm,bottom=1mm,
  title={ISS: Disputed Issues Identification},
  fonttitle=\bfseries\small,
  listing options={style=promptstyle}
}
### Task and Requirements
Task: I will provide the Reasoning section of a judgment. Extract a list of the disputed issues addressed in that section.

Definition: A disputed issue is an outcome-relevant factual or legal question that the court must resolve. It may concern rights and obligations between parties in a civil case, criminal liability or prosecution in a criminal case, or the legality of an administrative action in an administrative case; it need not be framed as an express disagreement between two private parties.

Requirements:
1. Extract only issues supported by the input text; do not infer or invent issues.
2. Use the finest reasonable granularity. Each item must express one indivisible, atomic disputed issue.
3. Preserve the legally material parties, acts, amounts, and conditions when they are necessary to identify an issue.
4. If no disputed issue is stated or can be directly extracted, return an empty list [].

### Output Format
Return a strict List[str], for example:
[
  "Whether the agreement reflects the parties' true intent and is legally valid",
  "Whether the extension of the construction period constitutes a breach by the defendant",
  "Whether the plaintiff's claim for CNY 200,000 in liquidated damages has a factual and legal basis"
]

### Input
{text}

Extract the results according to the requirements above:
\end{tcblisting}

\begin{tcblisting}{
  listing only,
  breakable,
  colback=black!2,
  colframe=black!45,
  boxrule=0.4pt,
  left=1mm,right=1mm,top=1mm,bottom=1mm,
  title={JUS: Judicial Justification},
  fonttitle=\bfseries\small,
  listing options={style=promptstyle}
}
### Task and Requirements
Task: I will provide the Reasoning section of a judgment. Extract a list of the court's key judicial justifications for all material conclusions, including the treatment of claims, determinations of criminal liability and sentencing, and administrative dispositions, as applicable.

Definition: A judicial justification is the court's analysis explaining why a particular legal conclusion or disposition follows from the established facts, disputed issues, and applicable law.

Requirements:
1. Extract only reasoning points supported by the input text; do not add new facts, legal grounds, or conclusions.
2. Use the finest reasonable granularity. Each item must express one indivisible, atomic reasoning point.
3. Retain the factual premise, legal basis, and resulting conclusion when they are explicitly connected in the text.
4. Do not treat a bare fact, statutory citation, or dispositive statement as a justification unless the text uses it as part of the court's reasoning.
5. If no judicial justification can be directly extracted, return an empty list [].

### Output Format
Return a strict List[str], for example:
["<reason 1>", "<reason 2>", "<reason 3>"]

### Input
{text}

Extract the results according to the requirements above:
\end{tcblisting}

\subsection{Criminal Case Dimensions}

\begin{tcblisting}{
  listing only,
  breakable,
  colback=black!2,
  colframe=black!45,
  boxrule=0.4pt,
  left=1mm,right=1mm,top=1mm,bottom=1mm,
  title={Charge: Criminal Charge Prediction},
  fonttitle=\bfseries\small,
  listing options={style=promptstyle}
}
### Task and Requirements
I will provide the Judgment section of a model-generated criminal judgment. Extract the criminal charge or charges of which each defendant is convicted.

Requirements:
1. Extract only criminal charges explicitly stated in the input text.
2. Use each defendant's name as a dictionary key and a list of that defendant's charge labels as the value.
3. Preserve charge labels in the source language.
4. Do not include charges stated only as allegations if the final judgment does not adopt them.
5. When a defendant is convicted of multiple charges, return each charge separately.
6. Do not include civil causes of action, administrative violations, sentencing factors, or punishment types.
7. If a defendant is identified but no charge of conviction can be extracted, return an empty list for that defendant.
8. If no defendant-specific charge information is present, return an empty dictionary {}.
9. Return only the required dictionary without any explanation.

### Output Format
Return a strict Python dictionary in the following form:
{"<defendant_name>": ["<charge_1>", "<charge_2>"]}

Example:
{"Defendant A": ["Charge 1", "Charge 2"], "Defendant B": []}

### Input
{text}

Extract the results according to the requirements above:
\end{tcblisting}

The extracted results are matched to the ground truth by defendant name. For each defendant in the ground truth, a charge appearing in both the extracted result and the ground truth is counted as a true positive. A charge appearing only in the extracted result is counted as a false positive, while a charge appearing only in the ground truth is counted as a false negative. Defendants appearing only in the extracted result are ignored.

For defendant \(i\) in case \(c\), let \(\mathrm{TP}_{c,i}\), \(\mathrm{FP}_{c,i}\), and \(\mathrm{FN}_{c,i}\) denote the corresponding numbers of true positives, false positives, and false negatives. The defendant-level Charge score is
\begin{equation}
F_{1,c,i}^{\mathrm{Charge}}
=
\frac{2\mathrm{TP}_{c,i}}
{2\mathrm{TP}_{c,i}+\mathrm{FP}_{c,i}+\mathrm{FN}_{c,i}}.
\label{eq:charge_defendant_score}
\end{equation}
If a defendant in the ground truth has no corresponding extracted result or has an empty extracted charge list, the defendant-level F$_1$ score is \(0\).

Let \(N_c\) denote the number of defendants in the ground truth for case \(c\). The case-level Charge score is
\begin{equation}
S_c^{\mathrm{Charge}}
=
\frac{1}{N_c}
\sum_{i=1}^{N_c}F_{1,c,i}^{\mathrm{Charge}}.
\label{eq:charge_case_score}
\end{equation}
Let \(K\) denote the number of evaluated criminal cases. The final Charge score is
\begin{equation}
\mathrm{Charge}
=
\frac{1}{K}
\sum_{c=1}^{K}S_c^{\mathrm{Charge}}.
\label{eq:charge_score}
\end{equation}

\begin{tcblisting}{
  listing only,
  breakable,
  colback=black!2,
  colframe=black!45,
  boxrule=0.4pt,
  left=1mm,right=1mm,top=1mm,bottom=1mm,
  title={Fine: Monetary Fine},
  fonttitle=\bfseries\small,
  listing options={style=promptstyle}
}
### Task and Requirements
I will provide the Judgment section of a criminal judgment. Extract the monetary fine imposed on each defendant.

Definition: A monetary fine is a criminal financial punishment imposed by the court. It does not include compensation, restitution, refunds, civil payment obligations, confiscation, litigation costs, or other monetary items.

Requirements:
1. Extract only fines explicitly imposed in the input text. Do not infer an amount.
2. Use each defendant's name as a dictionary key and the fine amount as the value.
3. Normalize every amount to an integer number of Chinese yuan and omit the currency unit and all other text.
4. If the judgment expressly states a final aggregate fine for a defendant, return that aggregate and do not double-count its component amounts.
5. If a defendant is mentioned but no fine is imposed, or a fine is mentioned without a numerical amount, return None for that defendant.
6. If no defendant-specific fine information is present, return an empty dictionary {}.

### Output Format
Return a strict Python dictionary in the following form:
{"<defendant_name>": <int or None>}

Example:
{"Defendant A": 5000, "Defendant B": None}

### Input
{text}

Extract the results according to the requirements above:
\end{tcblisting}

We apply the extraction prompt to the model-generated judgment to obtain structured fine predictions. The extracted results are matched to the ground truth by defendant name. Each defendant with a ground-truth fine is included in the evaluation. When a numerical predicted amount is available for the same defendant, the score is calculated using Equation~\ref{eq:NVS}. If the predicted value is missing, the corresponding defendant-level score is set to \(0\). Defendants appearing only in the prediction are ignored.

Let \(n_c\) denote the number of defendants with a ground-truth fine in case \(c\), and let \(s_{c,i}^{\mathrm{Fine}}\) denote the score for the \(i\)-th defendant. When a numerical prediction \(a_{c,i}\) is available, \(s_{c,i}^{\mathrm{Fine}}=\mathrm{NVS}(a_{c,i},b_{c,i})\), where \(b_{c,i}\) is the ground-truth amount. When the predicted value is missing, \(s_{c,i}^{\mathrm{Fine}}=0\). The case-level Fine score is
\begin{equation}
S_c^{\mathrm{Fine}}
=
\frac{1}{n_c}
\sum_{i=1}^{n_c}
s_{c,i}^{\mathrm{Fine}}.
\label{eq:fine_case_score}
\end{equation}
Let \(K\) denote the number of cases containing at least one defendant with a ground-truth fine. The final Fine score is
\begin{equation}
\mathrm{Fine}
=
\frac{1}{K}
\sum_{c=1}^{K}
S_c^{\mathrm{Fine}}.
\label{eq:fine_score}
\end{equation}

\begin{tcblisting}{
  listing only,
  breakable,
  colback=black!2,
  colframe=black!45,
  boxrule=0.4pt,
  left=1mm,right=1mm,top=1mm,bottom=1mm,
  title={REC: Recidivism Assessment},
  fonttitle=\bfseries\small,
  listing options={style=promptstyle}
}
### Task and Requirements
I will provide the Reasoning section of a model-generated criminal judgment. Determine, for each defendant, whether the court identifies that defendant as a recidivist.

Requirements:
1. Return True if the text explicitly identifies the defendant as a recidivist; otherwise return False.
2. Base the result only on the input text. Do not infer recidivism from prior convictions, repeated offending, or a criminal record unless the court explicitly states the legal determination.
3. Use each defendant's name as a dictionary key and a Boolean value as the result.
4. If no defendant is identifiable in the input, return an empty dictionary {}.
5. Return only the required dictionary without any explanation.

### Output Format
Return a strict Python dictionary in the following form:
{"<defendant_name>": <True or False>}

Example:
{"Defendant A": True, "Defendant B": False}

### Input
{text}

Extract the results according to the requirements above:
\end{tcblisting}

The extracted results are matched to the ground truth by defendant name. Each defendant in the ground truth is treated as one classification instance, and if no corresponding prediction is available, the predicted label for that defendant is set to \texttt{False}. Defendants appearing only in the extracted result are ignored.

The labels from all evaluated criminal cases are pooled before scoring, and no case-level averaging is performed. Let \(\mathrm{TP}\), \(\mathrm{FP}\), \(\mathrm{FN}\), and \(\mathrm{TN}\) denote the numbers of true positives, false positives, false negatives, and true negatives, respectively, with \texttt{True} representing the recidivist class. The F$_1$ score for the recidivist class is
\begin{equation}
F_1^{\mathrm{True}}
=
\frac{2\mathrm{TP}}
{2\mathrm{TP}+\mathrm{FP}+\mathrm{FN}},
\label{eq:rec_true_f1}
\end{equation}
and the F$_1$ score for the non-recidivist class is
\begin{equation}
F_1^{\mathrm{False}}
=
\frac{2\mathrm{TN}}
{2\mathrm{TN}+\mathrm{FP}+\mathrm{FN}}.
\label{eq:rec_false_f1}
\end{equation}
The final REC score is the macro-average of the two class-level scores:
\begin{equation}
\mathrm{REC}
=
\frac{F_1^{\mathrm{True}}+F_1^{\mathrm{False}}}{2}.
\label{eq:rec_score}
\end{equation}

\begin{tcblisting}{
listing only,
breakable,
colback=black!2,
colframe=black!45,
boxrule=0.4pt,
left=1mm,right=1mm,top=1mm,bottom=1mm,
title={MIT: Mitigating-Circumstance Identification},
fonttitle=\bfseries\small,
listing options={style=promptstyle}
}
### Task and Requirements
I will provide the Reasoning section of a model-generated criminal judgment. Determine, for each defendant, whether the court recognizes any sentencing circumstance that supports lenient treatment.

Requirements:
1. Return True if the text explicitly recognizes at least one circumstance supporting lenient treatment for the defendant; otherwise return False.
2. Such circumstances may include voluntary surrender, confession, remorse, restitution, compensation, or obtaining the victim's forgiveness.
3. Base the result only on the input text. Do not infer a mitigating circumstance from conduct that the court does not explicitly recognize as supporting lenient treatment.
4. Use each defendant's name as a dictionary key and a Boolean value as the result.
5. If no defendant is identifiable in the input, return an empty dictionary {}.
6. Return only the required dictionary without any explanation.

### Output Format
Return a strict Python dictionary in the following form:
{"<defendant_name>": <True or False>}

Example:
{"Defendant A": True, "Defendant B": False}

### Input
{text}

Extract the results according to the requirements above:
\end{tcblisting}

The extracted results are matched to the ground truth by defendant name. Each defendant in the ground truth is treated as one classification instance, and if no corresponding prediction is available, the predicted label for that defendant is set to \texttt{False}. Defendants appearing only in the extracted result are ignored.

The labels from all evaluated criminal cases are pooled before scoring, and no case-level averaging is performed. Let \(\mathrm{TP}\), \(\mathrm{FP}\), \(\mathrm{FN}\), and \(\mathrm{TN}\) denote the numbers of true positives, false positives, false negatives, and true negatives, respectively, with \texttt{True} indicating that at least one mitigating circumstance is recognized. The F$_1$ score for the positive class is
\begin{equation}
F_1^{\mathrm{True}}
=
\frac{2\mathrm{TP}}
{2\mathrm{TP}+\mathrm{FP}+\mathrm{FN}},
\label{eq:mit_true_f1}
\end{equation}
and the F$_1$ score for the negative class is
\begin{equation}
F_1^{\mathrm{False}}
=
\frac{2\mathrm{TN}}
{2\mathrm{TN}+\mathrm{FP}+\mathrm{FN}}.
\label{eq:mit_false_f1}
\end{equation}
The final MIT score is the macro-average of the two class-level scores:
\begin{equation}
\mathrm{MIT}
=
\frac{F_1^{\mathrm{True}}+F_1^{\mathrm{False}}}{2}.
\label{eq:mit_score}
\end{equation}

\begin{tcblisting}{
  listing only,
  breakable,
  colback=black!2,
  colframe=black!45,
  boxrule=0.4pt,
  left=1mm,right=1mm,top=1mm,bottom=1mm,
  title={Pen\_M: Penalty Duration Prediction},
  fonttitle=\bfseries\small,
  listing options={style=promptstyle}
}
### Task and Requirements
I will provide the sentencing portion of a criminal judgment. Extract the duration of the principal punishment imposed on each defendant and normalize it to months.

Requirements:
1. Extract the final duration of a duration-based principal punishment, including fixed-term imprisonment, criminal detention, or public surveillance.
2. Convert years and months to a total number of months. For example, three years is 36 months, and one year and six months is 18 months.
3. If both individual sentences and an expressly imposed combined sentence are stated, extract the duration of the combined principal punishment.
4. If the punishment is life imprisonment, return the string "LIFE_IMPRISONMENT".
5. If the punishment is the death penalty, including a death sentence with a reprieve, return the string "DEATH_PENALTY".
6. If the punishment is exemption from criminal punishment, or if no duration-based principal punishment is stated, return None.
7. If a suspended sentence is imposed for a duration-based principal punishment, extract the duration of the principal punishment and ignore the probation period.

### Output Format
Return a strict Python dictionary using each defendant's name as the key:
{"<defendant_name>": <integer_months or "LIFE_IMPRISONMENT" or "DEATH_PENALTY" or None>}

Example input:
Defendant Gao is sentenced to one year of fixed-term imprisonment, suspended for two years; Defendant Li is sentenced to life imprisonment; Defendant Wang is sentenced to the death penalty with a two-year reprieve.

Example output:
{"Defendant Gao": 12, "Defendant Li": "LIFE_IMPRISONMENT", "Defendant Wang": "DEATH_PENALTY"}

### Input
{text}

Extract the results according to the requirements above:
\end{tcblisting}

We apply the extraction prompt to the model-generated judgment to obtain structured punishment predictions. The extracted results are matched to the ground truth by defendant name. If the ground truth provides a numerical duration or a categorical punishment for a defendant but no corresponding punishment is extracted from the model-generated judgment, the defendant receives a score of \(0\) according to Equation~\ref{eq:penalty_instance_score}.

Let \(N_c\) denote the number of defendants for whom the ground truth provides such a punishment in case \(c\), and let \(s_{c,i}^{\mathrm{Pen\_M}}\) denote the score assigned to the \(i\)-th defendant according to Equation~\ref{eq:penalty_instance_score}. The case-level Pen\_M score is
\begin{equation}
S_c^{\mathrm{Pen\_M}}=\frac{1}{N_c}\sum_{i=1}^{N_c}s_{c,i}^{\mathrm{Pen\_M}},
\label{eq:penalty_case_score}
\end{equation}
where \(N_c>0\). Let \(K\) denote the number of cases with \(N_c>0\). The final Pen\_M score is
\begin{equation}
\mathrm{Pen\_M}=\frac{1}{K}\sum_{c=1}^{K}S_c^{\mathrm{Pen\_M}}.
\label{eq:penalty_score}
\end{equation}

\subsection{Civil Case Dimensions}

\begin{tcblisting}{
  listing only,
  breakable,
  colback=black!2,
  colframe=black!45,
  boxrule=0.4pt,
  left=1mm,right=1mm,top=1mm,bottom=1mm,
  title={SUP$_{\mathrm{Civ}}$: Civil Claim Support},
  fonttitle=\bfseries\small,
  listing options={style=promptstyle}
}
### Task and Requirements
I will provide the Reasoning and Judgment sections of a civil judgment. Determine the overall disposition of the plaintiff's claims.

Requirements:
1. Classify the overall disposition as exactly one of: "upheld", "partially upheld", or "rejected".
2. Return "upheld" when all claims decided by the court are fully upheld.
3. Return "rejected" when none of the claims decided by the court are upheld.
4. Return "partially upheld" when the plaintiff receives some but not complete support, including when a claim is only partially upheld or when multiple claims receive different outcomes.
5. Determine the outcome solely from the input text.
6. If the Reasoning and Judgment sections differ, follow the final operative Judgment section.
7. Return only the required dictionary without any explanation.

### Output Format
Return a strict Python dictionary in the following form:
{"outcome": "<upheld or partially upheld or rejected>"}

### Input
{text}

Determine the overall disposition according to the requirements above:
\end{tcblisting}

For each case, the predicted label is compared with the ground-truth label. For each outcome label, let \(P\) and \(R\) denote precision and recall, respectively. Its F$_1$ score is
\begin{equation}
F_1=\frac{2PR}{P+R}.
\label{eq:sup_civ_f1}
\end{equation}
In our evaluations, \(P+R>0\) for all three outcome labels. Let \(F_1^{U}\), \(F_1^{PU}\), and \(F_1^{R}\) denote the F$_1$ scores for \texttt{upheld}, \texttt{partially upheld}, and \texttt{rejected}, respectively. The final SUP$_{\mathrm{Civ}}$ score is their unweighted mean:
\begin{equation}
\mathrm{SUP}_{\mathrm{Civ}}
=
\frac{F_1^{U}+F_1^{PU}+F_1^{R}}{3}.
\label{eq:sup_civ_score}
\end{equation}

\begin{tcblisting}{
  listing only,
  breakable,
  colback=black!2,
  colframe=black!45,
  boxrule=0.4pt,
  left=1mm,right=1mm,top=1mm,bottom=1mm,
  title={PAY$_{\mathrm{Civ}}$: Civil Payment Obligation},
  fonttitle=\bfseries\small,
  listing options={style=promptstyle}
}
### Task and Requirements
I will provide the Judgment section of a civil judgment. Extract the total amount of civil payment obligations imposed on each obligor.

Requirements:
1. Extract only monetary payment obligations explicitly imposed in the final Judgment section.
2. Include compensation, repayment, damages, restitution, and other civil monetary liabilities.
3. Do not extract criminal fines, litigation costs, or non-monetary performance obligations.
4. Use the name of the person or entity bearing the payment obligation as the dictionary key.
5. If a final total is explicitly stated for an obligor, return that total. Otherwise, sum the explicitly stated payment amounts imposed on that obligor.
6. Normalize each amount to an integer number of Chinese yuan and omit the currency unit and all other text.
7. If an obligor is identified but no numerical amount can be extracted, return None for that obligor.
8. If no civil payment obligation is imposed, return an empty dictionary {}.
9. Return only the required dictionary without any explanation.

### Output Format
Return a strict Python dictionary in the following form:
{"<civil_obligor_name_or_entity>": <int or None>}

Example:
{"Company A": 5000, "Party B": None}

### Input
{text}

Extract the results according to the requirements above:
\end{tcblisting}

The extracted amounts are matched to the ground truth by obligor name. For each obligor with both a predicted amount and a ground-truth amount, the NVS score is calculated using Equation~\ref{eq:NVS}. If an obligor has a ground-truth amount but the predicted value is missing, the corresponding NVS score is set to \(0\). Obligors appearing only in the prediction are ignored. PAY is evaluated only on cases containing at least one reference payment obligation. Let \(n_c\) denote the number of obligors with a payment obligation in the reference judgment for civil case \(c\), and let \(s_{c,i}\) denote the score of the \(i\)-th obligor. The case-level score is
\begin{equation}
S_c^{\mathrm{PAY_{Civ}}}
=
\frac{1}{n_c}\sum_{i=1}^{n_c}s_{c,i}.
\label{eq:pay_civ_case_score}
\end{equation}

Let \(K\) denote the number of evaluated civil cases. The final PAY$_{\mathrm{Civ}}$ score is
\begin{equation}
\mathrm{PAY}_{\mathrm{Civ}}
=
\frac{1}{K}\sum_{c=1}^{K}S_c^{\mathrm{PAY_{Civ}}}.
\label{eq:pay_civ_score}
\end{equation}

\subsection{Administrative Case Dimensions}

\begin{tcblisting}{
  listing only,
  breakable,
  colback=black!2,
  colframe=black!45,
  boxrule=0.4pt,
  left=1mm,right=1mm,top=1mm,bottom=1mm,
  title={SUP$_{\mathrm{Adm}}$: Administrative Claim Support},
  fonttitle=\bfseries\small,
  listing options={style=promptstyle}
}
### Task and Requirements
I will provide the Reasoning and Judgment sections of an administrative judgment. Determine the overall disposition of the plaintiff's claims.

Requirements:
1. Classify the overall disposition as exactly one of: "upheld", "partially upheld", or "rejected".
2. Return "upheld" when all claims decided by the court are fully upheld.
3. Return "rejected" when none of the claims decided by the court are upheld.
4. Return "partially upheld" when the plaintiff receives some but not complete support, including when a claim is only partially upheld or when multiple claims receive different outcomes.
5. Classify the outcome of the plaintiff's claims, not whether the challenged administrative act remains in effect.
6. Determine the outcome solely from the input text.
7. If the Reasoning and Judgment sections differ, follow the final operative Judgment section.
8. Return only the required dictionary without any explanation.

### Output Format
Return a strict Python dictionary in the following form:
{"outcome": "<upheld or partially upheld or rejected>"}

### Input
{text}

Determine the overall disposition according to the requirements above:
\end{tcblisting}

For each case, the predicted label is compared with the ground-truth label. The label-level F$_1$ scores are calculated using Equation~\ref{eq:sup_civ_f1}. Let \(F_1^{U}\), \(F_1^{PU}\), and \(F_1^{R}\) denote the F$_1$ scores for \texttt{upheld}, \texttt{partially upheld}, and \texttt{rejected}, respectively. The final SUP$_{\mathrm{Adm}}$ score is their unweighted mean:
\begin{equation}
\mathrm{SUP}_{\mathrm{Adm}}
=
\frac{F_1^{U}+F_1^{PU}+F_1^{R}}{3}.
\label{eq:sup_adm_score}
\end{equation}

\begin{tcblisting}{
  listing only,
  breakable,
  colback=black!2,
  colframe=black!45,
  boxrule=0.4pt,
  left=1mm,right=1mm,top=1mm,bottom=1mm,
  title={PAY$_{\mathrm{Adm}}$: Administrative Payment Obligation},
  fonttitle=\bfseries\small,
  listing options={style=promptstyle}
}
### Task and Requirements
I will provide the Judgment section of an administrative judgment. Extract the total amount of administrative payment obligations imposed on each obligor.

Requirements:
1. Extract only monetary payment obligations explicitly imposed in the final Judgment section.
2. Include administrative compensation, restitution, refunds, and other monetary obligations ordered by the court.
3. Do not extract a monetary fine merely because it appears in the challenged administrative act.
4. Do not extract criminal fines, litigation costs, or non-monetary administrative dispositions.
5. Use the name of the person, entity, or administrative authority bearing the payment obligation as the dictionary key.
6. If a final total is explicitly stated for an obligor, return that total. Otherwise, sum the explicitly stated payment amounts imposed on that obligor.
7. Normalize each amount to an integer number of Chinese yuan and omit the currency unit and all other text.
8. If an obligor is identified but no numerical amount can be extracted, return None for that obligor.
9. If no administrative payment obligation is imposed, return an empty dictionary {}.
10. Return only the required dictionary without any explanation.

### Output Format
Return a strict Python dictionary in the following form:
{"<administrative_obligor_name_or_entity>": <int or None>}

Example:
{"Administrative Authority A": 5000, "Entity B": None}

### Input
{text}

Extract the results according to the requirements above:
\end{tcblisting}

The extracted amounts are matched to the ground truth by obligor name. For each obligor with both a predicted amount and a ground-truth amount, the NVS score is calculated using Equation~\ref{eq:NVS}. If an obligor has a ground-truth amount but the predicted value is missing, the corresponding NVS score is set to \(0\). Obligors appearing only in the prediction are ignored. PAY is evaluated only on cases containing at least one reference payment obligation. Let \(n_c\) denote the number of obligors with a payment obligation in the reference judgment for administrative case \(c\), and let \(s_{c,i}\) denote the score of the \(i\)-th obligor. The case-level score is
\begin{equation}
S_c^{\mathrm{PAY_{Adm}}}
=
\frac{1}{n_c}\sum_{i=1}^{n_c}s_{c,i}.
\label{eq:pay_adm_case_score}
\end{equation}
Let \(K\) denote the number of evaluated administrative cases. The final PAY$_{\mathrm{Adm}}$ score is
\begin{equation}
\mathrm{PAY}_{\mathrm{Adm}}
=
\frac{1}{K}\sum_{c=1}^{K}S_c^{\mathrm{PAY_{Adm}}}.
\label{eq:pay_adm_score}
\end{equation}

\begin{tcblisting}{
  listing only,
  breakable,
  colback=black!2,
  colframe=black!45,
  boxrule=0.4pt,
  left=1mm,right=1mm,top=1mm,bottom=1mm,
  title={DIS: Administrative Disposition Prediction},
  fonttitle=\bfseries\small,
  listing options={style=promptstyle}
}
### Task and Requirements
Task: I will provide the Judgment section of an administrative judgment. Extract a list of the core non-monetary administrative dispositions made by the court concerning the challenged administrative act or an administrative authority's statutory duties.

Definition: An administrative disposition is a judicial determination concerning the legal effect of a challenged administrative act or an administrative authority's ensuing non-monetary duties. It includes, but is not limited to, annulment or partial annulment of an administrative act, upholding an administrative act, declaring an administrative act unlawful or invalid, modifying an administrative act, ordering an administrative authority to perform a statutory duty, ordering an authority to issue a new administrative act, and other non-monetary dispositions with equivalent legal effect.

Requirements:
1. Extract only dispositions explicitly stated in the input text. Do not infer, supplement, or rewrite them based on the facts of the case.
2. Each list item must contain one indivisible, atomic disposition. If the text concerns multiple administrative acts, authorities, or dispositions, list them separately.
3. Retain enough information to identify each disposition, including the dispositive action, its object, and any expressly stated scope or condition, such as "in part", "within the statutory time limit", or "issue a new administrative act". Do not return context-free labels such as "annulled" or "upheld" alone.
4. If dismissal of the plaintiff's claims is the only explicit non-monetary outcome, extract that outcome as written; do not rewrite it as upholding the administrative act.
5. Do not extract administrative compensation, restitution, refunds, monetary fines, or other payment obligations under this dimension. A determination that annuls, upholds, or declares unlawful an administrative penalty decision involving a fine is nevertheless a non-monetary disposition and should be extracted.
6. Do not extract litigation-cost allocations, appeal deadlines, or other procedural information.
7. If no explicit non-monetary administrative disposition is present, return an empty list [].

### Output Format
Return a strict List[str], for example:
[
  "Annul Administrative Penalty Decision No. [X] issued by the defendant authority",
  "Order the defendant authority to issue a new administrative act within the statutory time limit"
]

### Input
{text}

Extract the results according to the requirements above:
\end{tcblisting}

\section{Semantic Matching Prompts and Deterministic Scoring}
\label{app:semantic_matching_prompts}

For each dimension \(d\in\{\mathrm{ISS},\mathrm{JUS},\mathrm{DIS}\}\), Qwen3-Plus receives the source case context and lists of model-output extracts and ground-truth items. In the templates below, \texttt{source\_case\_context} contains only the \textit{Header}, \textit{Background}, and \textit{Findings of Fact} provided to the judgment-generation model. Following the dimension-specific rubric, Qwen3-Plus assigns \textit{Equivalent}, \textit{Partially Equivalent}, or \textit{Not Equivalent} to every possible pair consisting of one model-output extract and one ground-truth item. These tags are mapped to fixed weights:

\begin{equation}
w(t)=
\begin{cases}
1, & t=\textit{Equivalent},\\
0.5, & t=\textit{Partially Equivalent},\\
0, & t=\textit{Not Equivalent}.
\end{cases}
\label{eq:semantic_tag_weights}
\end{equation}

For each case $c$, let $m_c$ and $n_c$ denote the numbers of ground-truth content items and model-output extracts, respectively. Let $T_i$ denote the $i$-th ground-truth content item and $M_j$ denote the $j$-th model-output extract. Let $t_{ij}^{(c)}$ denote the semantic tag assigned to the pair consisting of $T_i$ and $M_j$, and let $w(t_{ij}^{(c)})$ be its corresponding weight defined in Eq.~\ref{eq:semantic_tag_weights}. We then construct a complete bipartite graph between the $m_c$ ground-truth content items and the $n_c$ model-output extracts, where the edge between $T_i$ and $M_j$ has weight $w(t_{ij}^{(c)})$.

Let $\mathcal{M}_c$ denote the set of all valid one-to-one matchings between the ground-truth content items and the model-output extracts for case $c$, such that each item can participate in at most one matched pair. We define $W_{d,c}$ as the maximum total semantic weight achievable under such a matching:

\begin{equation}
W_{d,c} =
\max_{M \in \mathcal{M}_c}
\sum_{(i,j)\in M} w(t_{ij}^{(c)}).
\label{eq:maximum_matching_weight}
\end{equation}

The case-level soft semantic F$_1$ score is then defined as
\begin{equation}
S_{d,c}=
\begin{cases}
1, & m_c=n_c=0,\\
0, & \text{exactly one of } m_c,n_c \text{ is } 0,\\
\dfrac{2W_{d,c}}{m_c+n_c}, & \text{otherwise}.
\end{cases}
\label{eq:soft_semantic_f1}
\end{equation}

Let $N_d$ denote the number of applicable cases for dimension $d$. The final score for dimension $d$ is computed as the arithmetic mean of the case-level soft semantic F$_1$ scores:
\begin{equation}
\mathrm{Score}_d =
\frac{1}{N_d}
\sum_{c=1}^{N_d} S_{d,c}.
\label{eq:semantic_dimension_score}
\end{equation}

This case-level averaging gives equal weight to each applicable case, regardless of the numbers of ground-truth content items and model-output extracts in that case. The scoring method gives partial credit for partial semantic equivalence while penalizing both omitted ground-truth content and unsupported model-output extracts.

\subsection{ISS: Disputed Issues Identification}

\begin{tcblisting}{
  listing only,
  breakable,
  colback=black!2,
  colframe=black!45,
  boxrule=0.4pt,
  left=1mm,right=1mm,top=1mm,bottom=1mm,
  title={ISS: Pairwise Semantic Tagging},
  fonttitle=\bfseries\small,
  listing options={style=promptstyle}
}
### Task
Compare every ground-truth disputed issue with every model-output disputed issue. Each input item has a unique identifier. Assign exactly one semantic-equivalence tag to every possible pair consisting of one ground-truth disputed issue and one model-output disputed issue.

### Definition
A disputed issue is an outcome-relevant factual or legal question that the court must resolve. It may concern rights and obligations between parties in a civil case, a criminal accusation or defense in a criminal case, or the legality of an administrative act or omission in an administrative case. An express disagreement between two private parties is not required.

### Tags
- "Equivalent": The two items express the same outcome-relevant factual or legal question. Differences in wording, order, abstraction, or non-material detail do not change the question to be adjudicated.
- "Partially Equivalent": The two items substantially concern the same disputed question, but one is broader or narrower, or omits or changes a legally salient party, act, object, amount, condition, time period, or requested remedy.
- "Not Equivalent": The two items concern different factual or legal questions, contradict one another, or share only general case background without addressing the same adjudicative issue.

### Rules
1. Judge each pair independently at the atomic-issue level.
2. Compare legal meaning, not lexical overlap alone.
3. Do not tag two issues as Equivalent merely because they arise from the same case or may lead to the same outcome.
4. Use the source context only to resolve ambiguous parties, acts, objects, or references. Do not create an issue absent from either extracted item.
5. Output one comparison for every Cartesian-product pair. Do not omit or duplicate any pair.
6. Preserve the supplied item identifiers exactly.
7. If either item list is empty, return {"comparisons": []}.
8. Return only the specified JSON object, with no explanation or markdown fence.

### Output Schema
{
  "comparisons": [
    {
      "ground_truth_id": "T1",
      "model_output_id": "M1",
      "tag": "Equivalent | Partially Equivalent | Not Equivalent"
    }
  ]
}

### Source Case Context
{source_case_context}

### Ground-Truth Disputed Issues
{ground_truth_issues}

### Model-Output Disputed Issues
{model_output_issues}

### Output
\end{tcblisting}

\subsection{JUS: Judicial Justification}

\begin{tcblisting}{
  listing only,
  breakable,
  colback=black!2,
  colframe=black!45,
  boxrule=0.4pt,
  left=1mm,right=1mm,top=1mm,bottom=1mm,
  title={JUS: Pairwise Semantic Tagging},
  fonttitle=\bfseries\small,
  listing options={style=promptstyle}
}
### Task
Compare every ground-truth judicial-justification point with every model-output judicial-justification point. Each input item has a unique identifier. Assign exactly one semantic-equivalence tag to every possible pair consisting of one ground-truth judicial-justification point and one model-output judicial-justification point.

### Definition
A judicial-justification point is an atomic proposition in which the court connects a legally material fact, evidentiary assessment, or legal rule to an intermediate or final adjudicative conclusion. It explains why the court accepts or rejects a disputed proposition, claim, defense, charge, or disposition.

### Tags
- "Equivalent": The two items express the same material reasoning proposition. Their factual premise, legal standard or basis, inferential relationship, and resulting conclusion are legally compatible and substantively the same. Differences in wording, order, or non-material detail are permitted.
- "Partially Equivalent": The two items share a material premise, inferential basis, or legal conclusion, but one omits a salient fact, legal ground, analytical step, limitation, or conclusion, or contains a minor legally relevant deviation.
- "Not Equivalent": The two items rely on materially different or incompatible facts, legal grounds, reasoning paths, or conclusions. A shared fact, citation, or final outcome alone is insufficient.

### Rules
1. Judge each pair independently at the atomic-reason level.
2. Assess legal meaning and the fact-rule-conclusion relationship, not prose quality or lexical overlap.
3. Do not treat a bare shared fact, statutory citation, or disposition as sufficient for equivalence.
4. Do not tag an alternative but materially different rationale as Equivalent merely because it is independently plausible or reaches the same result.
5. Use the source context only to resolve ambiguity and case-specific references. Do not add reasoning absent from either extracted item.
6. Output one comparison for every Cartesian-product pair. Do not omit or duplicate any pair.
7. Preserve the supplied item identifiers exactly.
8. If either item list is empty, return {"comparisons": []}.
9. Return only the specified JSON object, with no explanation or markdown fence.

### Output Schema
{
  "comparisons": [
    {
      "ground_truth_id": "T1",
      "model_output_id": "M1",
      "tag": "Equivalent | Partially Equivalent | Not Equivalent"
    }
  ]
}

### Source Case Context
{source_case_context}

### Ground-Truth Judicial-Justification Points
{ground_truth_justification_points}

### Model-Output Judicial-Justification Points
{model_output_justification_points}

### Output
\end{tcblisting}

\subsection{DIS: Administrative Disposition Prediction}

\begin{tcblisting}{
  listing only,
  breakable,
  colback=black!2,
  colframe=black!45,
  boxrule=0.4pt,
  left=1mm,right=1mm,top=1mm,bottom=1mm,
  title={DIS: Pairwise Semantic Tagging},
  fonttitle=\bfseries\small,
  listing options={style=promptstyle}
}
### Task
Compare every ground-truth administrative disposition with every model-output administrative disposition. Each input item has a unique identifier. Assign exactly one semantic-equivalence tag to every possible pair consisting of one ground-truth administrative disposition and one model-output administrative disposition.

### Definition
An administrative disposition is a court's non-monetary determination concerning the legal effect of a challenged administrative act or an administrative authority's ensuing statutory duty. Examples include annulment or partial annulment, upholding an administrative act, declaring an act unlawful or invalid, modifying an act, ordering performance of a statutory duty, and ordering an authority to issue a new administrative act.

### Tags
- "Equivalent": The two dispositions have the same principal legal effect on the same administrative act, omission, authority, or statutory duty. Differences in wording, order, or non-material implementation detail are permitted.
- "Partially Equivalent": The dispositions have substantially overlapping legal effects, but one omits or changes a legally salient object, scope, condition, time limit, partial-versus-complete effect, or consequential order.
- "Not Equivalent": The dispositions concern different administrative acts or duties, contradict one another, or produce materially different legal effects, such as annulment versus upholding, performance versus dismissal, or unlawfulness versus validity.

### Rules
1. Judge each pair independently at the atomic-disposition level.
2. Compare the operative legal effect, its object, and any expressly stated scope or condition.
3. Do not treat dismissal of a plaintiff's claims as automatically equivalent to upholding the challenged administrative act.
4. Exclude monetary compensation, restitution, refunds, payment obligations, litigation costs, and appeal instructions. A ruling on the validity or legal effect of an administrative penalty decision involving a fine remains a non-monetary disposition.
5. Use the source context only to resolve ambiguous acts, authorities, duties, or references. Do not create a disposition absent from either extracted item.
6. Output one comparison for every Cartesian-product pair. Do not omit or duplicate any pair.
7. Preserve the supplied item identifiers exactly.
8. If either item list is empty, return {"comparisons": []}.
9. Return only the specified JSON object, with no explanation or markdown fence.

### Output Schema
{
  "comparisons": [
    {
      "ground_truth_id": "T1",
      "model_output_id": "M1",
      "tag": "Equivalent | Partially Equivalent | Not Equivalent"
    }
  ]
}

### Source Case Context
{source_case_context}

### Ground-Truth Administrative Dispositions
{ground_truth_dispositions}

### Model-Output Administrative Dispositions
{model_output_dispositions}

### Output
\end{tcblisting}

\section{Case Study}
\label{sec:case-study}

\begin{figure*}[ht]
\centering
\includegraphics[width=\textwidth]{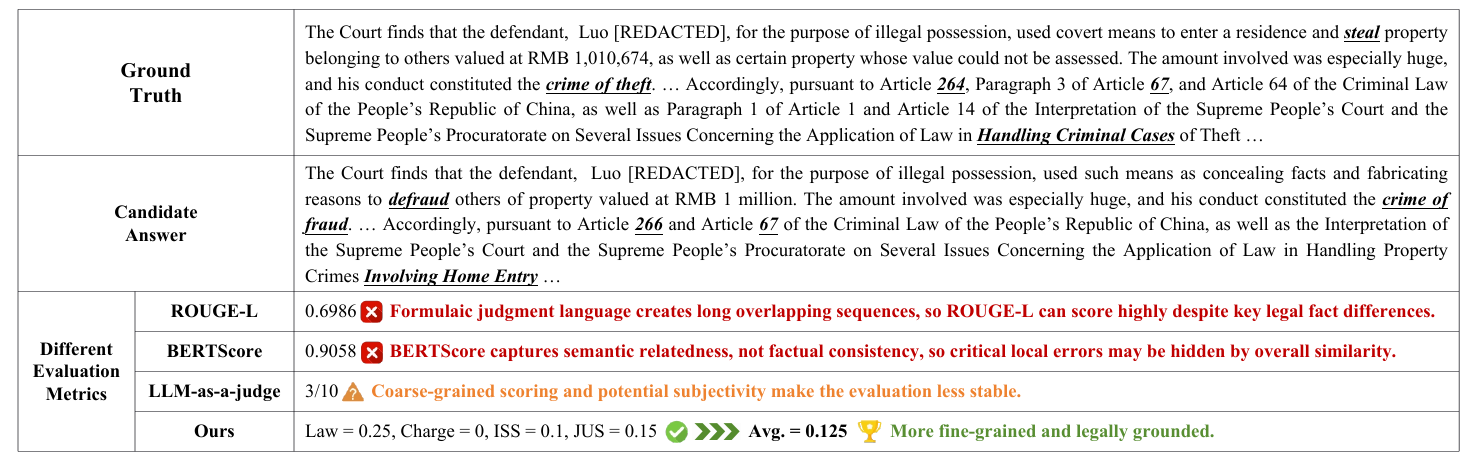}
\caption{
Case study comparing different evaluation metrics on a legally inconsistent judgment pair.
}
\label{fig:case-study}
\end{figure*}

As shown in Figure~\ref{fig:case-study}, this case study illustrates how conventional similarity-based metrics can overestimate consistency in legal judgment evaluation. Although the two texts share highly formulaic legal language, they differ in critical legal dimensions, including the charge, criminal conduct, statutory basis, and amount involved. ROUGE-L and BERTScore assign relatively high scores because they primarily measure lexical overlap or semantic similarity, while LLM-as-a-judge provides only a relatively coarse holistic assessment. In contrast, our method evaluates key legal dimensions separately and produces a much lower average score, better reflecting the substantial legal inconsistency between the two judgments.

\end{document}